\documentclass{article}

\usepackage[accepted]{icml2023}
\usepackage{microtype}
\usepackage{graphicx}
\usepackage{subfigure}
\usepackage{booktabs} 

\usepackage{wrapfig,lipsum,booktabs}
\usepackage{float}
\usepackage{graphicx}
\usepackage{subfigure}
\usepackage{pifont}

\usepackage{hyperref}
\hypersetup{hidelinks}

\usepackage[title]{appendix}

\usepackage{amsmath}
\usepackage{amssymb}
\usepackage{mathtools}
\usepackage{amsthm}

\usepackage{booktabs}
\usepackage{multirow}

\usepackage{mathrsfs}
\usepackage{makecell}

\usepackage[capitalize,noabbrev]{cleveref}

\usepackage{listings}
\usepackage{color}
\definecolor{dkgreen}{rgb}{0,0.6,0}
\definecolor{gray}{rgb}{0.5,0.5,0.5}
\definecolor{mauve}{rgb}{0.58,0,0.82}

\theoremstyle{plain}

\theoremstyle{definition}

\theoremstyle{remark}

\usepackage{pifont}
\usepackage{amsmath,amssymb}
\newcommand{\cmark}{\ding{51}}
\newcommand{\xmark}{\ding{55}}

\def\1{\bm{1}}

\def\rvc{{\mathbf{c}}}

\def\rvh{{\mathbf{h}}}

\def\rvm{{\mathbf{m}}}

\def\rvp{{\mathbf{p}}}

\def\rvx{{\mathbf{x}}}

\def\rvz{{\mathbf{z}}}

\def\rmB{{\mathbf{B}}}

\def\rmI{{\mathbf{I}}}

\def\rmW{{\mathbf{W}}}
\def\rmX{{\mathbf{X}}}

\DeclareMathAlphabet{\mathsfit}{\encodingdefault}{\sfdefault}{m}{sl}
\SetMathAlphabet{\mathsfit}{bold}{\encodingdefault}{\sfdefault}{bx}{n}

\def\gF{{\mathcal{F}}}

\def\gL{{\mathcal{L}}}
\def\gM{{\mathcal{M}}}
\def\gN{{\mathcal{N}}}

\def\sR{{\mathbb{R}}}

\newcommand{\E}{\mathbb{E}}

\usepackage{xcolor}

\newcommand{\improved}[1]
{{\color{black} #1}}
\newcommand{\old}[1]
{{\color{black} #1}}

\usepackage[textsize=tiny]{todonotes}

\usepackage[textsize=tiny]{todonotes}

\icmltitlerunning{Non-autoregressive Conditional Diffusion Models for Time Series Prediction}

\begin{document}

\twocolumn[
\icmltitle{Non-autoregressive Conditional Diffusion Models 
           for Time Series Prediction}

\icmlsetsymbol{equal}{*}

\begin{icmlauthorlist}
\icmlauthor{Lifeng Shen}{yyy}
\icmlauthor{James T. Kwok}{comp}
\end{icmlauthorlist}

\icmlaffiliation{yyy}{Division of Emerging Interdisciplinary Areas, The Hong Kong University of Science and Technology, Clear Water Bay, Hong Kong.}
\icmlaffiliation{comp}{Department of Computer Science and Engineering, The Hong Kong University of Science and Technology, Clear Water Bay, Hong Kong}
\icmlcorrespondingauthor{Lifeng Shen}{lshenae@connect.ust.hk}
\icmlcorrespondingauthor{James T. Kwok}{jamesk@cse.ust.hk}

\icmlkeywords{Machine Learning, ICML}

\vskip 0.3in
]

\printAffiliationsAndNotice{}  

\begin{abstract}
Recently, denoising diffusion models have led to significant breakthroughs in the generation of images, audio and text. However, it is still an open question on how to adapt their strong modeling ability to model time series. In this paper, we propose TimeDiff, a non-autoregressive diffusion model that achieves high-quality time series prediction with the introduction of two novel conditioning mechanisms: future mixup and autoregressive initialization. Similar to teacher forcing, future mixup allows parts of the ground-truth future predictions for conditioning, while autoregressive initialization helps better initialize the model with basic time series patterns such as short-term trends. Extensive experiments are performed on nine real-world datasets. Results show that TimeDiff consistently outperforms existing time series diffusion models, and also achieves the best overall performance across a variety of the existing strong baselines (including transformers and FiLM).
\end{abstract}

\section{Introduction}

Time series prediction has been applied to many real-world applications such as
economics~\cite{henrique2019literature}, transportation~\cite{4840324}, and energy~\cite{wang2011review}.
As time series prediction can be regarded as a conditional generation task, it is
natural to use diffusion models \cite{timegrad}.
By generating data through iterative denoising, diffusion models present a strong
ability for data generation and have led to breakthroughs in synthesizing
images \cite{rombach2022high},
audio \cite{chen2020wavegrad} and text \cite{DiffuSeq,NEURIPS2022DiffusionLM}.
However, it is still an open question on how to build a strong diffusion model for time series prediction.
A key challenge is that time series data usually involve complex dynamics, nonlinear patterns, and long temporal dependencies.
Time series prediction is thus challenging, especially when the prediction
horizon is long \cite{zhou2022film}.

Existing time-series diffusion models can be roughly divided into two categories
based on the decoding strategy.
The first one is
autoregressive  \cite{timegrad,yang2022diffusion},
in which future predictions
are generated one by one over time.
However, it has limited long-range prediction performance due to error accumulation and slow inference speed.
To alleviate these problems,
the second category
of diffusion models
is
non-autoregressive, such as
CSDI \cite{tashiro2021csdi} and SSSD
\cite{alcaraz2022diffusion}.
They perform conditioning in
the denoising networks'
intermediate layers
and introduce inductive bias
into the denoising objective.
However, as will be shown empirically in Section \ref{sec:exp_results},
their
long-range prediction
performance is still inferior to other time-series prediction models such as Fedformer~\cite{22fedformer} and NBeats~\cite{oreshkin2019n}.
This may be due to that their conditioning strategies are
borrowed from image or text data,
but not tailored for time series data.
Only using the denoising objective to introduce inductive bias may not be
sufficient to guide the conditioning network in capturing helpful information
from the lookback window, leading to
inaccurate predictions.

In this paper, we propose TimeDiff,
a conditional
non-autoregressive
diffusion model that is effective for long time series prediction.
Unlike CSDI and SSSD, it introduces additional inductive bias in the conditioning module that is tailor-made for time series.
Specifically, TimeDiff introduces two conditioning mechanisms:
(i) future mixup,
which randomly reveals parts of the ground-truth future predictions during
training, 
and
(ii) autoregressive initialization, which better initializes the model with
basic
components in the time series.
Experimental results on nine real-world datasets demonstrate superiority of 
TimeDiff
over existing time series diffusion models and
other recent strong baselines (such as time series transformers
\cite{nie2022time,22fedformer,wu2021autoformer,liu2021pyraformer,zhou2021informer}, DLinear~\cite{zeng2022transformers} and FiLM~\cite{zhou2022film}).

\section{Preliminaries}
\subsection{Diffusion Models}

Diffusion models consist of a forward diffusion process and a backward denoising process.
A well-known diffusion model is the denoising diffusion probabilistic model (DDPM) \cite{ddpm20}.
By gradually adding noise,
the forward diffusion process transforms an input $\rvx^0$ to a white Gaussian noise vector $\rvx^K$ in $K$ diffusion steps.
At step
$k\in[1,K]$,
the diffused sample $\rvx^{k}$ is obtained by
scaling $\rvx^{k-1}$ with $\sqrt{1-\beta_k}$  and
adding i.i.d. Gaussian noise,
as:
\begin{equation}
  q(\rvx^k|\rvx^{k-1})=\gN(\rvx^k; \sqrt{1-\beta_k}\rvx^{k-1},\beta_k\rmI),\notag
\end{equation}
where
$\beta_k\in[0, 1]$
is the noise variance
following a predefined schedule.
It can be shown that
\begin{equation}
q(\rvx^k|\rvx^{0})=\gN(\rvx^k;\sqrt{\bar{\alpha}_k}\rvx^{0},(1-\bar{\alpha}_k)\rmI),
\label{eq:x0_2_xk}
\end{equation}
where $\alpha_k:=1-\beta_k$ and $\bar{\alpha}_k:=\Pi_{s=1}^k\alpha_s$. Thus,
$\rvx^k$ can be directly obtained as
\begin{align}
\rvx^k=\sqrt{\bar{\alpha}_k}\rvx^{0}+\sqrt{1-\bar{\alpha}_k}\epsilon,
\label{eq:x0_samples_xk}
\end{align}
where $\epsilon$ is sampled from $\gN(\mathrm{0}, \rmI)$. Note that
(\ref{eq:x0_samples_xk}) also allows $\rvx^0$ to be easily recovered from $\rvx^k$.

The backward denoising process is a Markovian process:
\begin{equation}
  p_{\theta}(\rvx^{k-1}|\rvx^k)=\gN(\rvx^{k-1};\mu_{\theta}(\rvx^k,k),\Sigma_{\theta}(\rvx^k,k)).
\label{eq:denoise_k}
\end{equation}
In practice, $\Sigma_{\theta}(\rvx^k,k)$ is often fixed at $\sigma_k^2\rmI$, and
  $\mu_{\theta}(\rvx^k,k)$
is modeled
by a neural network parameterized by $\theta$.

To train the diffusion model, one uniformly samples $k$ from $\{1,2,\dots,K\}$
and then minimizes the  following
KL-divergence
\begin{align}
\gL_k = D_{\mathtt{KL}}\left(q(\rvx^{k-1}|\rvx^k)||p_{\theta}(\rvx^{k-1}|\rvx^k)\right).
\label{loss:old}
\end{align}
For
more stable training,
$q(\rvx^{k-1}|\rvx^k)$ is often replaced by
\begin{align}
q(\rvx^{k-1}|\rvx^k,\rvx^0)=\gN(\rvx^{k-1};\tilde{\mu}_k(\rvx^k, \rvx^0, k),
\tilde{\beta}_k\rmI),
\label{eq:x0_xk_2_xk1}
\end{align}
where
$\tilde{\beta}_k=\frac{1-\bar{\alpha}_{k-1}}{1-\bar{\alpha}_{k}}{\beta}_k$ and
\begin{align}
\tilde{\mu}_k(\rvx^k, \rvx^0, k)=\frac{\sqrt{\bar{\alpha}_{k-1}}\beta_k}{1-\bar{\alpha}_k}\rvx^0+\frac{\sqrt{\alpha_{k}}(1-\bar{\alpha}_{k-1})}{1-\bar{\alpha}_k}\rvx^k.\notag
\end{align}
The training objective in (\ref{loss:old}) is then transformed as
\begin{align}
\gL_k = \frac{1}{2\sigma_k^2}\|\tilde{\mu}_k(\rvx^k, \rvx^0, k)-{\mu}_{\theta}(\rvx^k, k)\|^2.
\label{loss:old2}
\end{align}

In (\ref{loss:old2}),
${\mu}_{\theta}(\rvx^k, k)$
can be defined
in two ways \cite{benny2022dynamic}:
(i) ${\mu}_{\epsilon}(\epsilon_{\theta})$ or (ii) ${\mu}_{\rvx}(\rvx_{\theta})$.
In the former case, ${\mu}_{\epsilon}(\epsilon_{\theta})$ is computed from a noise prediction model $\epsilon_{\theta}(\rvx^k, k)$:
\begin{align}
    {\mu}_{\epsilon}(\epsilon_{\theta}) = \frac{1}{\sqrt{\alpha_k}}{\rvx^k}-\frac{1-\alpha_k}{\sqrt{1-\bar{\alpha}_k}\sqrt{\alpha_k}}\epsilon_{\theta}(\rvx^k, k).
\label{eq:eps_network}
\end{align}
\citeauthor{ddpm20} (\citeyear{ddpm20}) show that optimizing the
following simplified training objective
leads to better generation quality:
\begin{equation}
  \gL_{\epsilon} = \E_{k, \rvx^0, \epsilon}\left[\|\epsilon-\epsilon_{\theta}(\rvx^{k},k)\|^2\right],
\label{eq:loss_eps}
\end{equation}
where $\epsilon$ is the noise
used to obtain $\rvx^k$ from $\rvx^0$ in
(\ref{eq:x0_samples_xk})
at step $k$.
Alternatively, ${\mu}_{\rvx}(\rvx_{\theta})$ can be obtained from a data prediction model ${\rvx}_{\theta}(\rvx^k, k)$ as
\begin{align}
    {\mu}_{\rvx}(\rvx_{\theta}) \!=\! \frac{\sqrt{\alpha_k}(1-\bar{\alpha}_{k-1})}{1-\bar{\alpha}_{k}}{\rvx^k}\!+\!\frac{\sqrt{\bar{\alpha}_{k-1}}\beta_k}{{1-\bar{\alpha}_k}}\rvx_{\theta}(\rvx^k, k).
\label{eq:x_0_network}
\end{align}
The
corresponding
simplified loss is
\begin{equation}
  \gL_{\rvx} = \E_{k, \rvx^0, \epsilon}\left[\|\rvx^{0}-\rvx_{\theta}(\rvx^{k},k)\|^2\right].
\label{eq:x_0_loss}
\end{equation}

Note that the noise prediction model $\epsilon_{\theta}(\rvx^k, k)$
in (\ref{eq:loss_eps})
and the data prediction model ${\rvx}_{\theta}(\rvx^k, k)$
in (\ref{eq:x_0_loss}) are both conditioned on the diffusion step $k$ only.
When
an additional condition input $\rvc$ is available,
this can be injected into the backward denoising step (\ref{eq:denoise_k}) as
\begin{equation}
  p_{\theta}(\rvx^{k-1}|\rvx^k,\rvc)=\gN(\rvx^{k-1};\mu_{\theta}(\rvx^k,k|\rvc),\sigma_k^2\rmI).
\label{eq:conditional_denoise_k}
\end{equation}
We
can then
obtain conditional extensions of (\ref{eq:eps_network})-(\ref{eq:x_0_loss}) by
replacing $\epsilon_{\theta}(\rvx^k, k)$ (resp. ${\rvx}_{\theta}(\rvx^k, k)$) with
$\epsilon_{\theta}(\rvx^k, k|\rvc)$ (resp. ${\rvx}_{\theta}(\rvx^k, k|\rvc)$).
For example, using the data prediction model, we have
\begin{align}
    {\mu}_{\rvx}(\rvx_{\theta}) = \frac{\sqrt{\alpha_k}(1-\bar{\alpha}_{k-1})}{1-\bar{\alpha}_{k}}{\rvx^k}+\frac{\sqrt{\bar{\alpha}_{k-1}}\beta_k}{{1-\bar{\alpha}_k}}\rvx_{\theta}(\rvx^k, k|\rvc).
\label{eq:cond_x_0_network}
\end{align}

\subsection{Conditional DDPMs for Time Series Prediction}

In time series prediction, we aim to predict the future values
$\rvx_{1:H}^0\in\sR^{d\times H}$ of a time series given its past observations
$\rvx_{-L+1:0}^0\in\sR^{d\times L}$. Here,
$d$ is the number of variables in the possibly multivariate time series, $H$ is the length of forecast window, and $L$ is the length of lookback window.

Conditional DDPMs perform time series prediction by modeling the
distribution
\begin{equation}
  p_{\theta}(\rvx_{1:H}^{0:K}|\rvc)\!=\!p_{\theta}(\rvx_{1:H}^{K})\prod_{k=1}^Kp_{\theta}(\rvx_{1:H}^{k-1}|\rvx_{1:H}^{k},\rvc),\notag
\end{equation}
where
$\rvx_{1:H}^{K}\sim\gN(\mathbf{0},\rmI)$,
$\rvc=\gF(\rvx_{-L+1:0}^0)$
is the output of the conditioning network $\gF$ that takes the past observations
$\rvx_{-L+1:0}^0$ as input, and
			
\begin{align}
p_{\theta}(\rvx_{1:H}^{k-1}|\rvx_{1:H}^{k}, \rvc)
=\gN(\rvx_{1:H}^{k-1};\mu_{\theta}(\rvx_{1:H}^{k},k|\rvc),\sigma_k^2\rmI). \notag
\end{align}
It is still an open question how to design an efficient denoising network $\mu_{\theta}$ and conditioning network $\gF$ in time series diffusion models.

TimeGrad~\cite{timegrad}
is a recent denoising diffusion model for time series prediction.
It generates future values in an autoregressive
manner by modeling the joint distribution $p_{\theta}(\rvx_{1:H}^{0:K})$,
where $\rvx_{1:H}^{0:K} =
\{\rvx_{1:H}^{0}\}\bigcup\{\rvx_{1:H}^{k}\}_{k=1,\dots,K}$:
\begin{eqnarray*}
\lefteqn{p_{\theta}\left(\rvx_{1:H}^{0:K}|\rvc=\gF(\rvx_{-L+1:0}^0)\right)} \notag \\
&  =&\prod_{t=1}^{H}p_{\theta}\left(\rvx_{t}^{0:K}|\rvc=\gF(\rvx_{-L+1:t-1}^{0})\right)  \notag \\
&  =& \prod_{t=1}^{H} p_{\theta}(\rvx_{t}^{K})\prod_{k=1}^Kp_{\theta}\left(\rvx_{t}^{k-1}|\rvx_{t}^{k}, \rvc=\gF(\rvx_{-L+1:t-1}^0)\right). \notag
\end{eqnarray*}
In TimeGrad,
$\gF$
is
a recurrent neural network
that uses its hidden state $\rvh_t$ as $\rvc$.
Similar to (\ref{eq:loss_eps}),
its training objective is
\begin{equation}
  \gL_{\epsilon} = \E_{k, \rvx^0, \epsilon}\left[\|\epsilon-\epsilon_{\theta}(\rvx^{k}_t, k|\rvh_t)\|^2\right].\notag
\end{equation}
TimeGrad has been successfully used for short-term time series
prediction. However,
with its use of
autoregressive decoding,
error can accumulate and inference is also slow.  These are particularly
problematic for long-range prediction.

Another
denoising diffusion model for time series prediction
is CSDI \cite{tashiro2021csdi}.
Instead of training a shared diffusion model across time steps as in
TimeGrad,
CSDI
avoids
autoregressive inference on future values
by diffusing and denoising the whole time series $\rvx_{-L+1:H}^0$.
During training,
the denoising model
takes $\rvx_{-L+1:H}^0$ and a binary mask $\rvm\in\{0,1\}^{d\times(L+H)}$ as input,
where $\rvm_{i,t}=0$ if position $i$ is observed at time $t$, and 1 otherwise.
A self-supervised strategy is  also introduced by further masking some
input observations.
The following loss is used during training:
\begin{equation}
  \gL_{\epsilon} = \E_{k, \rvx^0, \epsilon}\left[\|\epsilon-\epsilon_{\theta}(\rvx^{k}_{\mathtt{target}}, k|\rvc=\gF(\rvx^{k}_{\mathtt{observed}}))\|^2\right],\notag
\end{equation}
where $\rvx^{k}_{\mathtt{target}}
=\rvm\odot\rvx_{-L+1:H}^0$
is the masked part
of the time series,
and $\rvx^{k}_{\mathtt{observed}}
=(1-\rvm)\odot\rvx_{-L+1:H}^0$
is the observed part.

Although directly generating the future time series in this non-autoregressive
manner avoids the error accumulation issue in TimeGrad, CSDI is still limited in two aspects:
(i) CSDI's denoising network is based on two transformers, whose complexity is
quadratic in the number of variables and length of time series. This can easily
run out of memory when modeling long multivariate time series.  (ii) Its conditioning is based on masking, similar to inpainting in computer
vision. However, it is shown that this 				
may cause disharmony at the boundaries between masked and observed regions
\cite{lugmayr2022repaint}.

SSSD~\cite{alcaraz2022diffusion}
replaces the transformers in CSDI
by a structured state space model, thus
avoiding the quadratic complexity issue.
However, it still uses the same non-autoregressive strategy as CSDI, and can still
have deteriorated performance due
to
boundary disharmony.

There are some recent attempts in the NLP community to develop sequence diffusion models with non-autoregressive decoding over time, e.g., DiffuSeq \cite{DiffuSeq}.
Unlike natural language generation,
time series prediction is more challenging as this requires modeling
temporal dependencies on irregular, highly nonlinear, and noisy data.

\section{Proposed Model}

\begin{figure*}[t!]
\centering
\includegraphics[width=0.80\textwidth]{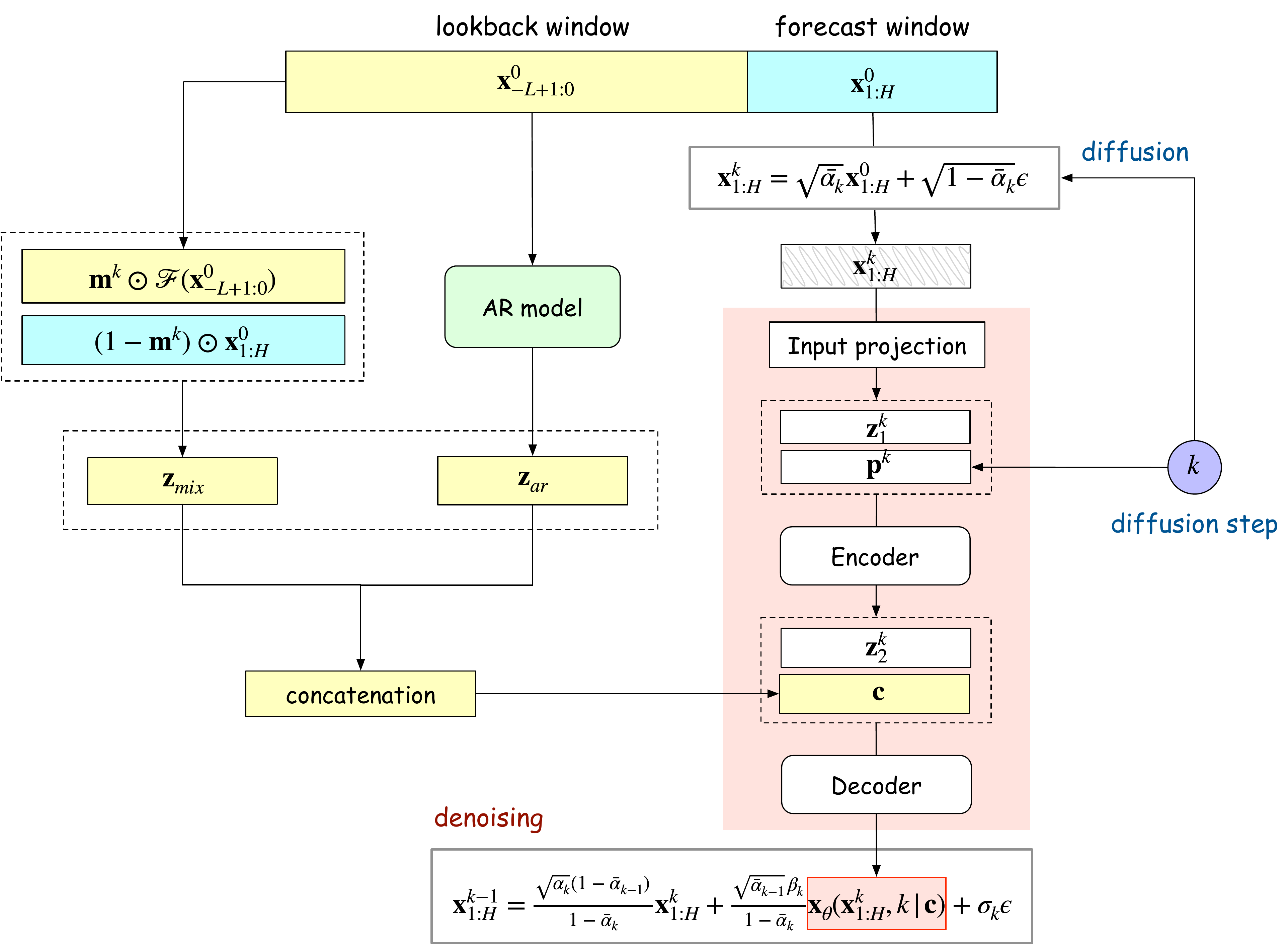}
\caption{An illustration of the proposed TimeDiff. $\rvx_{-L+1:0}^0$ contains the past observations and $\rvx_{1:H}^0$
contains the future ground-truth outputs. }
\label{fig:denoising_net}
\end{figure*}

While conditional diffusion models have been widely used,
they usually focus on capturing the semantic similarities across
modalities (e.g., text and image) ~\cite{choi2021ilvr,kim2022guided}.
However, to model
real-world non-stationary time series,
capturing the complex temporal
dependencies maybe even more important.
In this section, we propose TimeDiff, with novel conditioning mechanisms that are tailored for time series data.
Figure \ref{fig:denoising_net} shows an overview of the proposed
model.

\subsection{Forward Diffusion Process}

In TimeDiff, the forward diffusion process is  straightforward.
Based on (\ref{eq:x0_samples_xk}),
we obtain the diffused $\rvx_{1:H}^k$ by
$$
\rvx_{1:H}^k=\sqrt{\bar{\alpha}_k}\rvx_{1:H}^{0}+\sqrt{1-\bar{\alpha}_k}\epsilon,
$$
where  $\epsilon$ is sampled from $\gN(\mathrm{0}, \rmI)$ with the same size as $\rvx_{1:H}^{0}$.

The recent D$^3$VAE~\cite{ligenerative}
also uses the same
forward diffusion process on
the lookback window
$\rvx_{-L+1:0}^0$.
However,
strictly speaking,
D$^3$VAE
is not a diffusion model  as
the diffused $\rvx_{1:H}^k$
is produced
by a deep
variational auto-encoder (VAE)~\cite{kingma2013auto}
with $\rvx_{-L+1:0}^k$
(instead of $\rvx_{1:H}^k$)
as input
and does not denoise from random noise.
This makes the denoising process more difficult.

\subsection{Conditioning the Backward Denoising Process}
\label{sec:cond}

At the
$k$th denoising step, $\rvx_{1:H}^{k}$ is denoised to $\rvx_{1:H}^{k-1}$.
In order to well predict the future time series segment $\rvx_{1:H}^{0}$, useful
information needs to be extracted from the lookback window $\rvx_{-L+1:0}^0$ to guide the denoising of $\rvx_{1:H}^{k}$ to $\rvx_{1:H}^{0}$
(Figure \ref{fig:denoising_net}, left).

The proposed
inductive bias on the conditioning network
is specific to time series prediction.
Specifically,
Section~\ref{sec:future}
proposes the use of
mixup \cite{zhang2018mixup} to
combine
the past and future time series information into
$\rvz_{\mathtt{mix}}$ (in (\ref{eq:z_mix})).
Section~\ref{sec:ar}
proposes an autoregressive model to produce a crude approximation
$\rvz_{ar}$
(in (\ref{eq:ar})) of
$\rvx_{1:H}^{0}$.
Finally, these two are
concatenated
along the channel dimension
to form
the condition
\begin{align}
    \rvc = \text{concat}([\rvz_{\mathtt{mix}}, \rvz_{ar}]) \in
\sR^{2d\times H}.
\label{eq:cond}
\end{align}

\subsubsection{Future Mixup}								
\label{sec:future}

Since the goal is to predict
$\rvx_{1:H}^{0}$,
the ideal
condition
to guide the denoising process
is
$\rvx_{1:H}^{0}$ itself.
While obviously $\rvx_{1:H}^{0}$
cannot be accessed on inference, it is available during training.
Inspired by mixup~\cite{zhang2018mixup}, we propose
a simple yet effective mechanism called
{\em future mixup}.
It combines the past information's mapping $\gF(\rvx_{-L+1:0}^0)$ and the future
ground-truth $\rvx_{1:H}^{0}$. During training,
at diffusion step $k$, it produces the conditioning signal
\begin{align}
  \rvz_{\mathtt{mix}}=\rvm^k\odot\gF(\rvx_{-L+1:0}^0)+(1-\rvm^k)\odot\rvx_{1:H}^{0}.
\label{eq:z_mix}
\end{align}
Here,
$\rvm^k \in [0,1)^{d\times H}$ is a mixing matrix with each element randomly
sampled from
the uniform distribution on
$[0,1)$, and
$\odot$ is the Hadamard product.
We use a convolution network
as $\gF$,
which is commonly used for the
modeling of local temporal patterns and long-range dependencies
in time series \cite{wang2017time,li2019enhancing}.
On inference,
$\rvx_{1:H}^{0}$ is no longer available, and
we simply
set
\begin{align}
  \rvz_{\mathtt{mix}}=\gF(\rvx_{-L+1:0}^0).
\label{eq:z_mix_test}
\end{align}

Future mixup is similar to teacher forcing
\cite{williams1989teacherforcing} and scheduled sampling
\cite{bengio2015scheduled}, which also introduce ground-truth observation as input
during training but only use the model's prediction during inference.
However,
future mixup is for non-autoregressive conditional generation in time series diffusion models, while teacher forcing and scheduled sampling are for autoregressive decoding of recurrent networks.
Moreover, future mixup mixes the
past observations'
embedding
and future time
series (which is also similar to the bidirectional LSTM~\cite{graves2013hybrid}),
while
teacher forcing and
scheduled sampling
replace the model's prediction at the previous step by
the ground-truth past observation.

\subsubsection{Autoregressive Model}
\label{sec:ar}

In image inpainting applications,
non-autoregressive models
often produce disharmony at the boundaries between masked and observed regions
\cite{lugmayr2022repaint}. In the context of time series prediction, this
translates to
disharmony between the history and forecast segments, as will also be
empirically demonstrated in Section~\ref{sec:exp_results}.

To alleviate this problem, we
use a linear autoregressive (AR) model $\gM_{ar}$ to
provide
an initial guess $\rvz_{ar}\in\sR^{d\times H}$
for $\rvx_{1:H}^0$.
Let $\rvx_{i}^0$ be the $i$th column of $\rvx_{-L+1:0}^0$. We define
\begin{align}
\rvz_{ar} = \sum_{i=-L+1}^{0}\rmW_i\odot\rmX_{i}^0+\rmB,
\label{eq:ar}
\end{align}
where
$\rmX_{i}^0\in\sR^{d\times H}$ is a matrix containing $H$ copies of $\rvx_{i}^0$,
and $\rmW_i$'s $\in\sR^{d\times H}$, $\rmB \in\sR^{d\times H}$
are trainable parameters. 				

The AR model
$\gM_{ar}$ is pretrained on the training set
by minimizing
the $\ell_2$-distance between $\rvz_{ar}$ and the
ground-truth $\rvx_{1:H}^0$.  			
The number of pretraining epochs can be few
(in the experiments,
it is set to 20).

While obviously this simple AR model cannot
accurately approximate a complex nonlinear time series in general,
it can still capture simple patterns, such as short-term trends
\cite{lai2018modeling}.
Moreover, note that though $\gM_{ar}$ is an autoregressive model, it does not require
autoregressive decoding. In other words,
all columns of
$\rvz_{ar}$
are obtained
 simultaneously, instead of obtaining them one by one sequentially. This
 avoids the error accumulation and slow inference problems in TimeGrad.

\begin{algorithm}[t!]
\caption{Training.}\label{alg:training}
\begin{algorithmic}[1] 
\REQUIRE
Number of diffusion steps $K$; pretrained AR model $\gM_{ar}$.
\REPEAT
    \STATE Sample $\rvx_{1:H}^0$ from the training set;
    \STATE $k\sim \texttt{Uniform}(\{1,2,\dots,K\})$;
    \STATE ${\epsilon} \sim \mathcal{N}(\mathbf{0}, \mathbf{I})$;
    \STATE generate diffused sample $\rvx_{1:H}^k$ (as in (\ref{eq:x0_samples_xk})) by\\
    $\rvx_{1:H}^k=\sqrt{\bar{\alpha}_k}\rvx_{1:H}^{0}\!+\!\sqrt{1-\bar{\alpha}_k}\epsilon$;
    \STATE obtain diffusion step $k$'s embedding $\rvp^k$ using (\ref{eq:pk});
    \STATE randomly generate a matrix $\rvm^k$ in (\ref{eq:z_mix});
    \STATE obtain $\rvz_{\mathtt{mix}}$ by \emph{future mixup} using (\ref{eq:z_mix});
    \STATE obtain $\rvz_{ar}$ by (\ref{eq:ar});
    \STATE obtain condition $\rvc$ based on $\rvz_{\mathtt{mix}}$ and $\rvz_{ar}$
	 by (\ref{eq:cond});
    \STATE use the denoising network to generate denoised sample ${\rvx}_{1:H}^{k-1}$ by (\ref{eq:denoise_kth});
    \STATE calculate the loss $\gL_k(\theta)$ in (\ref{eq:loss_k});
    \STATE take gradient descent step on $\nabla_{\theta}L_k(\theta)$;
\UNTIL {converged}.
\end{algorithmic}
\end{algorithm}

\begin{algorithm}[t!]
\caption{Inference.}\label{alg:samping}
\begin{algorithmic}[1] 
\REQUIRE
Trained denoising network $\rvx_{\theta}$; trained
conditioning network $\gF$; pretrained AR model $\gM_{ar}$.
\STATE ${\rvx}^K_{1:H} \sim \mathcal{N}(\mathbf{0}, \mathbf{I})$;
\FOR {$k=K,\dots,1$}
    \STATE ${\epsilon} \sim \mathcal{N}(\mathbf{0}, \mathbf{I})$, if $k>1$, else
	 $\epsilon=0$;
    \STATE obtain diffusion step $k$'s embedding $\rvp^k$ using (\ref{eq:pk});
    \STATE obtain $\rvz_{\mathtt{mix}}$ by (\ref{eq:z_mix_test});
    \STATE obtain $\rvz_{ar}$ by (\ref{eq:ar});
    \STATE obtain condition $\rvc$ based on $\rvz_{\mathtt{mix}}$ and $\rvz_{ar}$
	 using (\ref{eq:cond});
    \STATE sample $\hat{\rvx}_{1:H}^{k-1}$ by (\ref{eq:denoise_kth});
    \STATE ${\rvx}_{1:H}^{k-1}=\hat{\rvx}_{1:H}^{k-1}$;
\ENDFOR
\RETURN $\hat{\rvx}_{1:H}^0$.
\end{algorithmic}
\end{algorithm}

\subsection{Denoising Network}\label{sec:denoising}

The denoising network is shown in red
in Figure \ref{fig:denoising_net}. 		
When denoising $\rvx_{1:H}^k\in\sR^{d\times H}$, the diffusion-step embedding $\rvp^k$
is first combined with the diffused input $\rvx_{1:H}^{k}$'s
embedding
$\rvz_1^k\in\sR^{d'\times H}$, where $\rvz_1^k$ is obtained by an input projection block consisting of several convolution layers.
As in \cite{timegrad,tashiro2021csdi,kong2020diffwave}, we obtain
the representation
$\rvp^k\in\sR^{d'}$
of diffusion step $k$
using
the transformer's sinusoidal position embedding \cite{vaswani2017attention}.
Specifically, we first have
\begin{align}
k_{\mathtt{embedding}} = [\sin(10^{\frac{0\times4}{w-1}}t),\dots,\sin(10^{\frac{w\times4}{w-1}}t),\notag\\ \cos(10^{\frac{0\times4}{w-1}}t),
    \dots,\cos(10^{\frac{w\times4}{w-1}}t) ],\notag
\end{align}
where $w=\frac{d'}{2}$,
and then use two fully-connected (FC) layers both with default hidden dimensions of 128
on $k_{\mathtt{embedding}}$ to
obtain
\begin{align}
    \rvp^k = \mathrm{SiLU}(\mathrm{FC}(\mathrm{SiLU}(\mathrm{FC}(k_{\mathtt{embedding}}))))
\in\sR^{d'\times 1},
\label{eq:pk}
\end{align}
where $\mathrm{SiLU}$ is the sigmoid-weighted linear unit ~\cite{silu}.

For the concatenation, $\rvp^k$ is broadcasted
over length to form
$[\rvp^k, \dots, \rvp^k]\in\sR^{d'\times H}$,
and then concatenated with $\rvz_1^k$ along the channel dimension.
The concatenated result is a tensor of size $2d'\times H$.
A multilayer convolution-based
encoder is then used to obtain the representation $\rvz_2^k\in\sR^{d''\times H}$.

A decoder is used to fuse $\rvc$ and $\rvz_2^k$.
First, we concatenate $\rvc$ and $\rvz_2^k$ along the variable
dimension to
generate an input of size $(2d+d'')\times H$.
The
decoder consists of multiple convolution layers.
Its output $\rvx_{\theta}(\rvx_{1:H}^k, k|\rvc)$
is of size $d \times H$, the same as $\rvx_{1:H}^k$.
Finally,
we generate the denoised output $\hat{\rvx}_{1:H}^{k-1}$ as
in (\ref{eq:cond_x_0_network}):
\begin{align}
\hat{\rvx}_{1:H}^{k-1} =&
\frac{\sqrt{\alpha_k}(1\!-\!\bar{\alpha}_{k\!-\!1})}{1\!-\!\bar{\alpha}_{k}}{\rvx_{1:H}^k}\!+\!\frac{\sqrt{\bar{\alpha}_{k-1}}\beta_k}{{1\!-\!\bar{\alpha}_k}}\rvx_{\theta}(\rvx_{1:H}^k,
k|\rvc) \nonumber\\
 &+\sigma_{k} \epsilon,
\label{eq:denoise_kth}
\end{align}
where ${\epsilon} \sim \mathcal{N}(\mathbf{0}, \mathbf{I})$.

Note that we
predict
the data
${\rvx}_{\theta}(\rvx_{1:H}^k, k)$
for denoising, rather than
predicting
the noise
$\epsilon_{\theta}(\rvx_{1:H}^k, k)$.
As time series data usually contain highly irregular noisy components,
estimating the diffusion noise
$\epsilon$ can be more difficult.

\subsection{Training} \label{seq:training}
The training procedure is shown  in
Algorithm \ref{alg:training}.
For each
$\rvx_{1:H}^0$,
we first randomly sample a batch of diffusion steps $k$'s, and
then minimize
a conditioned variant of
(\ref{eq:x_0_loss}):
$\min_{\theta}\gL(\theta)=\min_{\theta}\E_{\rvx_{1:H}^0,\epsilon\sim\gN(\mathbf{0},\rmI),k}\gL_k(\theta)$,
where
\begin{align}
\gL_k(\theta) = \|\rvx_{1:H}^0-{\rvx}_{\theta}(\rvx^k_{1:H}, k|\rvc)\|^2.
\label{eq:loss_k}
\end{align} 															

\subsection{Inference}
During inference
(Algorithm \ref{alg:samping}),
we first generate a noise vector
${\rvx}_{1:H}^K\sim\mathcal{N}(\mathbf{0}, \mathbf{I})$ of size $d\times H$. By
repeatedly running the denoising step (\ref{eq:denoise_kth}) till $k$ equals 1 	
(${\epsilon}$ is set to zero when $k=1$), we obtain the time series $\hat{\rvx}_{1:H}^0$ as final prediction.

\begin{table}[t!]
\setlength{\tabcolsep}{3pt}
\centering
\footnotesize
\caption{Summary of dataset statistics, including dimension, total observations, sampling frequency, and prediction length used in the experiments.}
\begin{tabular}{ccccccc}%
\midrule[1pt]
dataset & dim &  \#observations  & freq. & $H$ (steps) \\\midrule
{\it NorPool} & 18  & 70,128 &  1 hour & 1 month (720) \\
\textit{Caiso} & 10& 74,472  &  1 hour & 1 month (720)  \\
\textit{Weather} & 21 & 52,696  & 10 mins & 1 week (672)\\
\textit{ETTm1} & 7  & 69,680  & 15 mins & 2 days (192) \\
\textit{Wind} & 7 & 48,673  & 15 mins & 2 days (192) \\
\textit{Traffic} & 862 & 17,544 & 1 hour & 1 week (168) \\
\textit{Electricity} & 321 & 26,304  & 1 hour & 1 week (168) \\
\textit{ETTh1} & 7  & 17,420  & 1 hour & 1 week (168)\\
\textit{Exchange} & 8  & 7,588  & 1 day & 2 weeks (14) \\
\midrule[1pt]
\end{tabular}
\label{tab:data}
\end{table}

\begin{table*}[t!]
\centering
\caption{
Testing MSE
in the univariate setting.
Number in brackets is the rank.
The best is in bold, while the second best is underlined.}
\renewcommand\arraystretch{1.2}
\resizebox{.98\textwidth}{!}{
\begin{tabular}{c|lllllllll|c}%
\midrule[1pt]
\multirow{1}[0]{*}{} & \textit{NorPool}& \textit{Caiso}& \textit{Weather}& \textit{ETTm1} & \textit{Wind}  & \textit{Traffic}& \textit{Electricity}& \textit{ETTh1}& \textit{Exchange}& avg rank \\\midrule
\textbf{TimeDiff}& \underline{0.636} (2) & 0.122 (3)& \textbf{0.002} (2)& \underline{0.040} (2)& 2.407 (9) & \textbf{0.121} (1) & \textbf{0.232} (1) & \textbf{0.066} (1)& \underline{0.017} (3)&  \textbf{2.7} \\\midrule
TimeGrad& 1.129 (15)& 0.325 (15)& \textbf{0.002} (2)& 0.048 (6.5)& 2.530 (12)& {1.223} (16)& {0.920} (16)& 0.078 (8)& 0.041 (13.5)& 11.6 \\
CSDI& {0.967} (14)& {0.192} (9)& \textbf{0.002} (2)& {0.050} (10)& {2.434} (10.5)& {0.393} (13)& {0.520} (12)& \improved{0.083} (11)& {0.071} (16)& 10.8\\
SSSD& 1.145 (16)& \old{0.176} (7)& 0.004 (6)& 0.049 (8.5)& 3.149 (15)& 0.151 (6)& 0.370 (5)& 0.097 (14)& 0.023 (10.5)& 9.8\\
D$^3$VAE& 0.964 (13)& 0.521 (16)& \underline{0.003} (4.5)& 0.044 (4)& 2.679 (13)& 0.151 (6)& 0.535 (13)& 0.078  (8)& 0.019 (7)& 9.4\\
\midrule
FiLM&0.707 (5)& 0.185 (8)& 0.007 (10)& \textbf{0.038} (1)& \textbf{2.143} (1)& 0.198 (10)& 0.260 (3)& \underline{0.070} (2.5)& 0.018 (5.5)& 5.1 \\
Depts& 0.668 (3)& \textbf{0.107} (1)& 0.024 (13)& 0.046 (5)& 3.457 (16)& 0.151 (6)& 0.380 (8)& \underline{0.070} (2.5)& 0.020 (8.5)& 7.0 \\
NBeats& 0.768 (6)& 0.125 (4)& 0.137 (15)& 0.048 (6.5)& 2.434 (10.5) & 0.142 (4)& 0.378 (7)& 0.095 (13)& \textbf{0.016} (1)& 7.4\\
\midrule
PatchTST& \textbf{0.595} (1)& 0.193 (10)& 0.026 (14)& 0.052 (12)& 2.698 (14)& 0.177 (9)&0.450 (11)& 0.106 (15)& 0.020 (8.5)& 10.5\\
FedFormer& 0.891 (9)& 0.164 (5)& 0.005 (7.5)& 0.065 (15)& 2.351 (8)& 0.173 (8)& 0.376 (6)& 0.076 (5)& 0.050 (15)& 8.7\\
Autoformer& 0.946 (12)&0.248 (11)& \underline{0.003} (4.5)& 0.051 (11)& 2.349 (7)& 0.473 (14)& 0.659 (15)& 0.081 (10)& 0.041 (13.5)& 10.9\\
Pyraformer& 0.933 (11)& 0.165 (6)& 0.020 (12)& {0.054}{} (13) & 2.279 (3)& \underline{0.136} (2)&0.389 (9)& 0.076 (5)& \underline{0.017} (3)& 7.1\\
Informer& 0.804 (7)& 0.250 (12.5)& 0.007 (10)& 0.049 (8.5)& 2.297 (4)& 0.213 (11)& 0.363 (4)&0.076 (5)& 0.023 (10.5)& 8.1\\
Transformer& 0.928 (10)& 0.250 (12.5) & 0.007 (10)& 0.058 (14)& 2.306 (6)& 0.238 (12)& 0.430 (10)& 0.092 (12)& 0.018 (5.5)& 10.2\\
\midrule
DLinear& 0.671 (4)& \underline{0.118} (2)& 0.168 (16)&  0.041 (3)& \underline{2.171} (2)& {0.139} (3)& \underline{0.244} (2)& 0.078 (8)& \underline{0.017} (3) & \underline{4.8} \\
LSTMa& 0.836 (8)& 0.253 (14)& 0.005 (7.5)& 0.091 (16)& 2.299 (5)& 1.032 (15)& 0.596 (14)& 0.167 (16)& 0.031 (12)& 11.9\\
\midrule[1pt]
\end{tabular}
}
\label{tab:UNI_MSE}
\end{table*}

\section{Experiments}
\label{sec:expt}

In this section, we perform extensive experiments to compare the proposed TimeDiff
with a variety of
time series prediction models
on nine real-world  datasets.

\subsection{Setup}
\label{sec:setup}

Experiments are performed on nine real-world time series
datasets
(Table~\ref{tab:data})
\cite{zhou2021informer,wu2021autoformer,fan2022depts}: 
(i) {\it NorPool}\footnote{\url{https://www.nordpoolgroup.com/Market-data1/Power-system-data}},
which includes
eight years of hourly energy production volume series in multiple European countries;
(ii) {\it Caiso}\footnote{\url{http://www.energyonline.com/Data}},
which contains eight years of hourly actual electricity load series in different zones of California; (iii) {\it Traffic}\footnote{\url{http://pems.dot.ca.gov}},
which records the hourly road occupancy rates generated by
sensors in the San Francisco Bay area
freeways;
(iv) {\it Electricity}\footnote{\url{https://archive.ics.uci.edu/ml/datasets/ElectricityLoadDiagrams20112014}},
which includes the hourly electricity consumption of 321 clients over two years;
(v) {\it Weather}\footnote{\url{https://www.bgc-jena.mpg.de/wetter/}},
which records 21 meteorological indicators at 10-minute intervals from 2020 to 2021;
(vi) {\it Exchange}\cite{lai2018modeling}, which  describes the daily exchange rates of eight countries
(Australia, British, Canada, Switzerland, China, Japan, New Zealand, and
Singapore);
(vii)-(viii) {\it ETTh1} and {\it ETTm1}, which contain two years of electricity transformer temperature data \cite{zhou2021informer} collected in China, at 1-hour and 15-minute intervals, respectively;
(ix) {\it Wind}, which contains wind power records from 2020-2021 at 15-minute intervals \cite{ligenerative}.

For \textit{NorPool}, following
\cite{fan2022depts},
the training set covers observations before April 1, 2020,
the validation set is from April 1 to October 1, 2020, while
the test set is for after October 1, 2020.
For \textit{Caiso}, following
\cite{fan2022depts},
the training set covers observations before January 1, 2020,
the validation set is for January 1 to October 1, 2020, while
the test set is for after October 1, 2020.
For the other datasets, we follow \cite{wu2021autoformer,22fedformer} and
split the whole data dataset into training, validation, and test sets in
chronological order with the ratio of 6:2:2 for
\textit{ETTh1} and \textit{ETTm1},
and 7:1:2 for \textit{Weather}, \textit{Wind}, \textit{Traffic},
\textit{Electricity}, and \textit{Exchange}.

\begin{table*}[t!]
\centering 
\caption{
Testing MSE
in the multivariate setting.
Number in brackets is the rank.
The best is in bold, while the second best is underlined.
CSDI
runs out of memory on
\textit{Traffic} and \textit{Electricity}.
}
\renewcommand\arraystretch{1.2}
\resizebox{.98\textwidth}{!}{
\begin{tabular}{c|lllllllll|c}%
\midrule[1pt]
\multirow{1}[0]{*}{} & \textit{NorPool}& \textit{Caiso}& \textit{Weather}& \textit{ETTm1} & \textit{Wind}  & \textit{Traffic}& \textit{Electricity}& \textit{ETTh1}& \textit{Exchange}& avg rank  \\\midrule
\textbf{TimeDiff} & \underline{0.665} (2) & \underline{0.136}  (2)&\textbf{0.311}(1)& \textbf{0.336} (1)& \textbf{0.896} (1)& 0.564 (3)& \textbf{0.193} (1) & \textbf{0.407} (1)& \underline{0.018} (3)& \textbf{1.7} \\\midrule
TimeGrad& {1.152} (15)& {0.258} (14)& {0.392}{} (10)& {0.874}{} (14)&  {1.209} (15)& {1.745} (15)& {0.736} (15)& {0.993} (15)& {0.079} (13)& 13.9\\
CSDI& {1.011} (14) & {0.253} (13)& {0.356} (5)& {0.529} (11)& {1.066} (5)  & -& -& {0.497} (4)& {0.077} (12)& 10.6 \\
SSSD& {0.872}{} (8)& {0.195}{} (6)& {0.349}{} (4)& 0.464 (9)& 1.188 (13)& {0.642}{} (6)& {0.255}{} (7)&  0.726 (12)&0.061 (9) & 8.1\\
D$^3$VAE& {0.745}{} (5)& {0.241}{} (12) & {0.375} {} (7)& 0.362 (4)&  1.118 (11)& {0.928}{} (12)& \old{0.286}{} (10)& 0.504 (5)&  {0.200} (15)& 8.9\\
\midrule
FiLM& 0.723 (4)& 0.179 (4)&\underline{0.327} (2)&0.347 (3)& 0.984 (3)& 0.628 (5)&0.210 (3)&0.426 (3)& \textbf{0.016} (1.5)& \underline{3.2} \\
Depts& \textbf{0.662} (1)& \textbf{0.106} (1)& 0.761 (14)& 0.380 (6)&  1.082 (8)& 1.019 (14)&0.319{} (12)& 0.579 (9.5)& 0.020 (4)& 7.7\\
NBeats& 0.832 (6)&0.141 (3)& 1.344 (16)& 0.391 (7)& 1.069 (6)& \textbf{0.373} (1)& 0.269 (8)& 0.586 (11)& \textbf{0.016} (1.5)& 6.6\\
\midrule
PatchTST& {0.851}{} (7)& {0.193}{} (5)& 0.782 (15)& {0.372} (5)& {1.070} (7)& {0.831} (11)& {0.225}{} (5)& 0.526 (7)& 0.047 (7)& 7.7 \\
FedFormer& {0.873} (9)& {0.205} (7)& {0.342} (3)& 0.426 (8)&  {1.113} (10)& {0.591} (4)&  {0.238} (6)& {0.541} (8)& 0.133 (14)& 7.6\\
Autoformer& 0.940 (10)& 0.226 (10)& 0.360 (6)& 0.565 (12)& 1.083 (9)& 0.688 (10)& \underline{0.201} (2)& 0.516 (6)& 0.056 (8)& 9.0\\
Pyraformer& 1.008 (13)& 0.273 (15)& 0.394 (11)& {0.493} (10)& {1.061} (4) & {0.659} (7)& {0.273} (9)& {0.579} (9.5)& 0.032 (6)& 9.4\\
Informer& 0.985 (11)& 0.231 (11)& 0.385 (8)& 0.673 (13)& 1.168 (12)& 0.664 (8)& 0.298 (11)& 0.775 (14)& 0.073 (11) & 10.9\\
Transformer& 1.005 (12) & 0.206 (8)& 0.388 (9)& 0.992 (15)& 1.201 (14) & 0.671 (9)& 0.328 (13)& 0.759 (13)& 0.062 (10)& 11.3\\
\midrule
DLinear& 0.670 (3)& 0.461 (16)& 0.488 (12)& \underline{0.345} (2)& \underline{0.899} (2)& \underline{0.389} (2)& 0.215 (4)& \underline{0.415} (2)& 0.022 (5)& 5.3 \\
LSTMa& 1.481 (16)& 0.217 (9)& 0.662 (13)& 1.030 (16)& 1.464 (16)& 0.966 (13)& 0.414 (14)&  1.149 (16)& 0.403 (16)& 14.2\\
\midrule[1pt]
\end{tabular}
}
\label{tab:MUL_MSE}
\end{table*}

As can be seen in
Table \ref{tab:data},
all the datasets used are multivariate.
Besides running the models directly on these
multivariate datasets, we also convert them to
univariate time series for performance comparison.
For \textit{NorPool} and \textit{Caiso},
the univariate time series
are extracted
from all variables as in \cite{fan2022depts}.
For the other datasets, we follow \cite{wu2021autoformer,22fedformer} and extract the
univariate time series by
using the last variable
only.
Further details and example visualizations are in Appendix \ref{vis:raw_data}.

A variety of baselines are
compared, including:
(i) Time series diffusion models, including TimeGrad~\cite{timegrad},
CSDI~\cite{tashiro2021csdi}, SSSD~\cite{alcaraz2022diffusion}, and
D$^3$VAE~\cite{ligenerative};
(ii) SOTA
time series prediction
methods,
including FiLM~\cite{zhou2022film}, Depts~\cite{fan2022depts} and NBeats~\cite{oreshkin2019n};
(iii) Time series transformers, including PatchTST
\footnote{PatchTST uses a channel-independence setup, in which each
variate of the multivariate time series is predicted independently. However, this
then cannot assess the model's ability to capture multivariate dependencies and
may not be fair to the other models.
Thus, we follow the traditional setup
and does not use
channel-independence
in PatchTST. Additional experiments on using the channel-independence setup can be
found in Appendix \ref{appendix:channel_independence}.
}
\cite{nie2022time},
Fedformer~\cite{22fedformer},
Autoformer~\cite{wu2021autoformer}, Pyraformer~\cite{liu2021pyraformer},
Informer~\cite{zhou2021informer} and the standard
Transformer~\cite{vaswani2017attention}; and
(iv) DLinear~\cite{zeng2022transformers} and
LSTMa \cite{bahdanau2015neural},
an attention-based LSTM~\cite{hochreiter1997long}.
For all methods, the history length is selected from \{96, 192, 720, 1440\} by using the validation set.

For the proposed TimeDiff, the data is preprocessed
with
instance normalization~\cite{kim2021reversible} as in
\cite{zhou2022film}.
Specifically,
we subtract
the mean value in the lookback window
from each time series variable, and then divide it by the lookback window's
standard deviation. On inference, the mean and standard deviation are added back to the final prediction.
For the transformer baselines,
series stationarization~\cite{liu2022non}
is
used
as in \cite{liu2022non}.

\begin{figure*}[t!]
\centering
\subfigure[CSDI. \label{fig:csdi}]
{\includegraphics[width=0.33\textwidth]{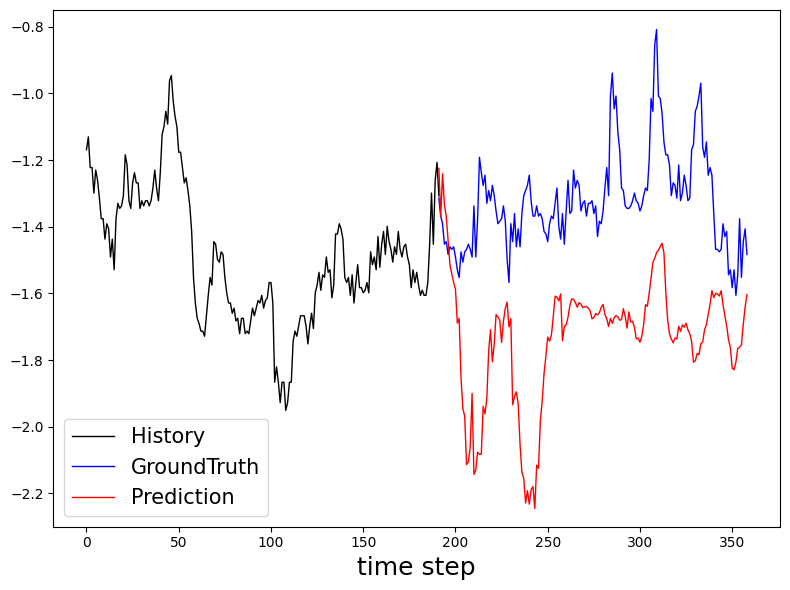}}
\subfigure[SSSD. \label{fig:sssd}]
{\includegraphics[width=0.33\textwidth]{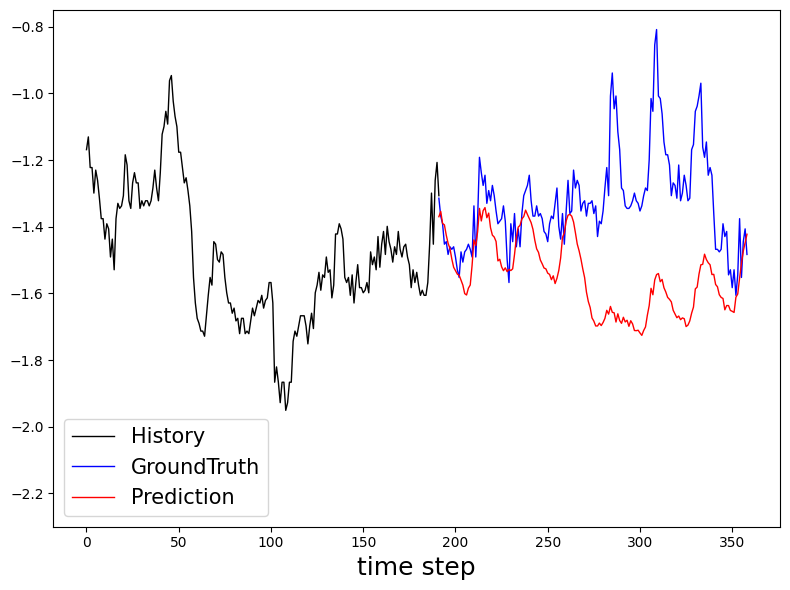}}
\subfigure[TimeDiff. \label{fig:ours}]
{\includegraphics[width=0.33\textwidth]{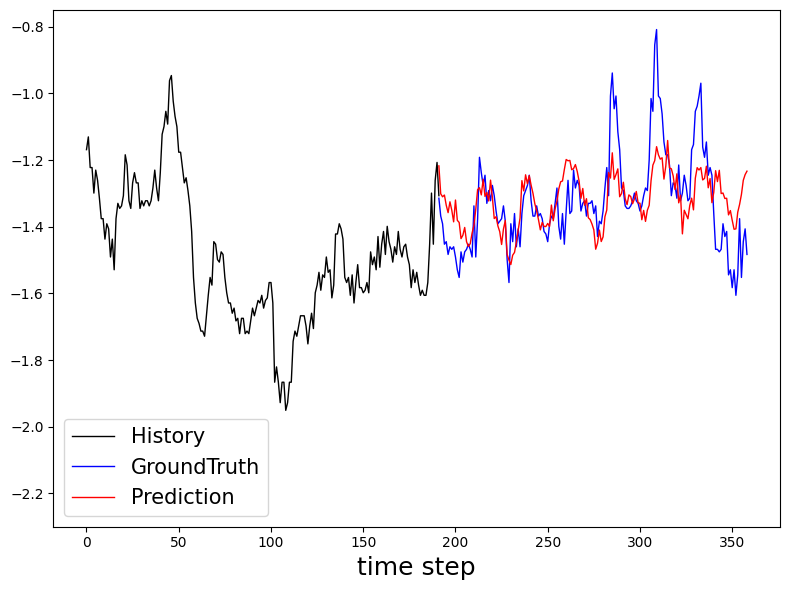}}
\caption{Visualizations on \textit{ETTh1} by CSDI, SSSD and the proposed TimeDiff.}
\label{fig:etth1_vis_baselines}
\end{figure*}

We train the proposed model using Adam \cite{kingma2014adam}
with a learning rate of $10^{-3}$.
The batch size is 64, and training
with early stopping for
a maximum of 100 epochs.
$K=100$
diffusion steps  are used,
with a cosine variance
schedule~\cite{timegrad}
starting from
$\beta_1=10^{-4}$ to $\beta_K=10^{-1}$.
To accelerate the inference of diffusion models,
many learning-free efficient samplers have been developed, such as
DDIM~\cite{song2021denoising}, Analytic-DPM~\cite{bao2022analyticdpm}, and
DPM-Solver~\cite{lu2022dpmsolver}. In this work, we use the
DPM-Solver.
Empirically, the number of denoising steps can be reduced to below
20.

The mean squared error (MSE)
is used
for performance evaluation.
As in \cite{timegrad,tashiro2021csdi},
the time series diffusion models
are evaluated
by averaging over 10 predictions.
All experiments are run on a Nvidia RTX A6000 GPU with 40GB memory.  More implementation details can be found in Appendix \ref{appendix:Implementation}.

\subsection{Results}\label{sec:exp_results}

Tables \ref{tab:UNI_MSE} and \ref{tab:MUL_MSE} show
the testing MSEs in the univariate and multivariate settings.
As can be seen, the proposed method
consistently outperforms existing time series diffusion models
(TimeGrad,
CSDI,
SSSD,
and
D$^3$VAE).
It also
achieves the best overall performance across all the baselines.
Note that CSDI
can easily run out of memory on long multivariate time series (Table \ref{tab:MUL_MSE})
due to its use of transformers.

Figure \ref{fig:etth1_vis_baselines} compares the predictions
obtained by the three non-autoregressive diffusion models CSDI, SSSD, and the
proposed TimeDiff
on a randomly selected
(univariate) \textit{ETTh1}
sample.
As can be seen,
while CSDI and SSSD
provide good
prediction in the
very short-term
(from steps 192-200),
their longer-term predictions differ significantly  from the ground-truth.
On the other hand, TimeDiff better captures both the trend and periodic
patterns.

\subsection{Ablation Studies}

In this section, we study the effectiveness of the
various proposed components.
Four representative univariate datasets in Table~\ref{tab:data} are used:
\textit{Caiso} and \textit{Electricity},
which contain obvious periodic patterns;
and \textit{Exchange} and \textit{ETTh1},
which are nonstationary (as discussed
in Appendix \ref{vis:raw_data}).

\subsubsection{Conditioning mechanism}

In this experiment, we study the effectiveness of  future mixup (section~\ref{sec:future}) and autoregressive model (AR) (section~\ref{sec:ar}).
We consider the four combinations when each of these
is used or not used in the conditioning network.
When neither is used, we set the condition
$\rvc$
to
$\gF(\rvx_{-L+1:0}^0)$.

Table \ref{tab:abl_conditioning} shows
the testing MSE.
As can be seen, both future mixup and AR model lead to improved performance.
In particular,
the performance
on \textit{ETTh1} degrades significantly when future mixup is not used.

\begin{table}[t!]
\centering
\caption{MSEs by different variants of
the conditioning network.}
\resizebox{.48\textwidth}{!}{
\begin{tabular}{cc|cccccccc}
\midrule[1pt]
\multirow{1}[0]{*}{future mixup} & \multirow{1}[0]{*}{AR} &  \multicolumn{1}{c}{\textit{Caiso}} &  \multicolumn{1}{c}{\textit{Electricity}}  &   \multicolumn{1}{c}{\textit{Exchange}} &   \multicolumn{1}{c}{\textit{ETTh1}}\\ \midrule
\cmark & \cmark &  \textbf{0.122} & \textbf{0.232} & \textbf{0.017} & \textbf{0.066}\\
\cmark & \xmark & 0.149 & 0.328 & 0.020 & 0.086\\
\xmark & \cmark & 0.160 & 0.295 & 0.022 & 0.162\\
\xmark & \xmark & 0.170 & 0.340 & 0.024 & 0.182\\
\midrule[1pt]
\end{tabular}
}
\label{tab:abl_conditioning}
\end{table}

\subsubsection{Mixup Strategies in Future Mixup}

Recall that in
future mixup,
elements of $\rvm^k$
are randomly
sampled from the uniform distribution     on
$[0,1)$.
In this experiment, we compare this strategy (which will be called soft-mixup)
with
two variants:
(i) Hard-mixup:
The sampled values in
$\rvm^k$
are binarized
by a threshold $\tau\in(0,1)$								
and (ii) Segment-mixup: The mask
$\rvm^k$
is generated by the procedure in \cite{transformer_repr}. Each masked
segment has a length following the geometric distribution with a mean of $3$. This is
then
followed by
an unmasked segment with mean length $3(1-\tau)/\tau$.

Results are shown in Table \ref{tab:abl_masking}.
Note that soft-mixup does not require the extra hyper-parameter $\tau$.
As can be seen, soft-mixup
has the best prediction performance. On the other hand,
hard-mixup
is sensitive to the setting of $\tau$.

\begin{table}[h!]
\centering
\caption{MSEs of different mixing strategies in future mixup. }
\resizebox{.48\textwidth}{!}{
\begin{tabular}{c|c|cccc}
\midrule[1pt]
 & $\tau$ &\multicolumn{1}{c}{\textit{Caiso}}&\multicolumn{1}{c}{\textit{Electricity}} & \multicolumn{1}{c}{\textit{Exchange}}& \multicolumn{1}{c}{\textit{ETTh1}}\\ \midrule
soft-mixup  & - & {0.122} & \textbf{0.232} & \textbf{0.017} & \textbf{0.066}\\\midrule
\multirow{5}[0]{*}{hard-mixup}
&  0.1 & 0.146 & 0.289 & 0.031 & 0.161\\
&  0.3 & 0.123 & 0.318 & 0.021 & 0.143\\
&  0.5 & \textbf{0.118} & 0.260 & 0.018 & 0.081\\
&  0.7 & 0.122 & 0.255 & 0.017 & 0.074\\
& 0.9 & 0.124 & 0.264  & 0.018 & 0.074 \\\midrule
\multirow{5}[0]{*}{segment-mixup}
&  0.1  & 0.152 &  0.335 & 0.032 & 0.172\\
&  0.3 & 0.121 & 0.323  & 0.020 & 0.132\\
&  0.5  & 0.119 & 0.285 & 0.018 & 0.079\\
&  0.7 &  0.124 & 0.254 & 0.018 & 0.072\\
&  0.9  & 0.125 & 0.253 & 0.018 & 0.075\\
\midrule[1pt]
\end{tabular}
}
\label{tab:abl_masking}
\end{table}

\subsubsection{Predicting ${\rvx}_{\theta}$ vs Predicting $\epsilon_{\theta}$}

In this experiment, we compare with the more common denoising strategy that is
based on
noise prediction
${\mu}_{\epsilon}(\epsilon_{\theta})$
\cite{timegrad}.
Here, the architecture of noise prediction network $\epsilon_{\theta}$ is the same
as that of the denoising network ${\rvx}_{\theta}$ in previous sections. The only
difference is that they have different objectives: predicting noise or predicting data.

Results are shown in Table \ref{tab:abl_data_or_noise}.
As can be seen,
predicting
the data
${\rvx}_{\theta}$ directly
is more effective.
This might be due to that real-world time series usually contains highly nonlinear noise,
which can be easily confused with the noise generated from the diffusion process.

\begin{table}[t!]
\centering
\caption{MSEs of two denoising strategies: Predicting ${\rvx}_{\theta}$ vs predicting $\epsilon_{\theta}$.}
\resizebox{.48\textwidth}{!}{
\begin{tabular}{c|cccc}
\midrule[1pt]
 denoising strategy & \multicolumn{1}{c}{\textit{Caiso}} & \multicolumn{1}{c}{\textit{Electricity}} & \multicolumn{1}{c}{\textit{Exchange}}&  \multicolumn{1}{c}{\textit{ETTh1}}\\ \midrule
 ${\rvx}_{\theta}$ & \textbf{0.122} & \textbf{0.232} & \textbf{0.017} & \textbf{0.66}  \\
  $\epsilon_{\theta}$ & 0.134 & 0.317 & 0.021 & 0.077   \\
\midrule
\end{tabular}
}
\label{tab:abl_data_or_noise}
\end{table}

\subsection{Integration into Existing Diffusion Models}
The proposed future mixup and autoregressive initialization are general techniques
and can be used with existing non-autoregressive time series diffusion models.
In this section, we integrate them into two non-autoregressive diffusion models: CSDI
and SSSD.
Specifically, we concatenate the  conditioning network's outputs with their conditioning masks along the channel dimension.
Experiment is performed on
the univariate \textit{ETTh1} and \textit{ETTm1}.

Table \ref{tab:improved_csdi_sssd} shows the results.
As can be seen, using future mixup and autoregressive initialization leads to
improved performance. 
Note, however, that even with these enhancements, CSDI still falls short of the
performance achieved by the proposed TimeDiff. Moreover, while the enhanced
variant of SSSD achieves comparable performance as TimeDiff on \textit{ETTm1}, it
is still outperformed by TimeDiff on \textit{ETTh1}.

\begin{table}[h!]
\centering
\caption{MSEs of
CSDI and  SSSD with and without
future mixup / autoregressive (AR) initialization.}
\resizebox{.40\textwidth}{!}{
\begin{tabular}{ccc|cc}
\midrule[1pt]
& future mixup & AR & \multicolumn{1}{c}{\textit{ETTh1}} & \multicolumn{1}{c}{\textit{ETTm1}} \\ \midrule
\multirow{4}[0]{*}{CSDI}
 & \xmark & \xmark & {0.083} & {0.050}  \\
 & \cmark & \xmark & 0.078 & 0.045 \\
 & \xmark & \cmark & 0.088 & 0.054 \\
 & \cmark & \cmark & \textbf{0.075} & \textbf{0.043} \\\midrule
\multirow{4}[0]{*}{SSSD}
 & \xmark & \xmark & {0.097} & {0.049}  \\
 & \cmark & \xmark & 0.077 & 0.044 \\
 & \xmark & \cmark & 0.084 & 0.052 \\
 & \cmark & \cmark & \textbf{0.071} & \textbf{0.040} \\\midrule
\end{tabular}
}
\label{tab:improved_csdi_sssd}
\end{table}

\subsection{Inference Efficiency}

In this experiment, we
compare the inference efficiency of the proposed TimeDiff with the other time
series diffusion model baselines (TimeGrad, CSDI, SSSD).
Table \ref{tab:runtime} shows the inference time 
on the univariate \textit{ETTh1} with different values of the prediction horizon $H$. 
As can be seen, the proposed TimeDiff is significantly faster than the others
across all the $H$ values.
In particular,
TimeGrad is the slowest
due to its use of auto-regressive decoding. 

\begin{table}[h!]
\centering
\caption{Inference time (ms) 
on the univariate \textit{ETTh1} with different prediction horizon $H$.} 
\resizebox{.48\textwidth}{!}{
\begin{tabular}{c|cccc}
\midrule[1pt]
& $H=96$ & $H=168$ & $H=336$ & $H=720$ \\ \midrule
TimeDiff & \textbf{16.2} & \textbf{17.2} & \textbf{26.5} & \textbf{34.6} \\\midrule
TimeGrad&	870.2&	1579.2	&3119.7	&6724.1\\
CSDI&	90.41	&127.2	&398.9	&513.1\\
 SSSD & 418.6	&595.0&	1054.2	&2516.9\\
\midrule
\end{tabular}
}
\label{tab:runtime}
\end{table}

\section{Conclusion}

In this paper, we propose TimeDiff, a novel diffusion model for time series prediction.
By using two conditioning mechanisms (future mixup and autoregressive initialization),
useful inductive bias is added to the conditioning network's outputs
and helps the denoising process.
Results on a number of real-world datasets show that the proposed method produces
better prediction results than existing time series diffusion models. The
proposed method
also achieves
the best overall performance across
existing strong baselines.
Besides, ablation studies demonstrate the effectiveness of each component in the proposed model.

One limitation of TimeDiff is that it has
difficulties in learning the multivariate dependencies when the time series has a large number of variables (e.g., \textit{Traffic}).
To alleviate this problem, in the future we will consider
capturing the dependencies
by integrating with graph neural networks.

\section*{Acknowledgements}
This research was supported in part by the Research Grants Council
of the Hong Kong Special Administrative Region (Grant
16200021).

\bibliography{main}

\begin{thebibliography}{48}
\providecommand{\natexlab}[1]{#1}
\providecommand{\url}[1]{\texttt{#1}}
\expandafter\ifx\csname urlstyle\endcsname\relax
  \providecommand{\doi}[1]{doi: #1}\else
  \providecommand{\doi}{doi: \begingroup \urlstyle{rm}\Url}\fi

\bibitem[Alcaraz \& Strodthoff(2022)Alcaraz and
  Strodthoff]{alcaraz2022diffusion}
Alcaraz, J. M.~L. and Strodthoff, N.
\newblock Diffusion-based time series imputation and forecasting with
  structured state space models.
\newblock Technical report, arXiv, 2022.

\bibitem[Bahdanau et~al.(2015)Bahdanau, Cho, and Bengio]{bahdanau2015neural}
Bahdanau, D., Cho, K., and Bengio, Y.
\newblock {Neural machine translation by jointly learning to align and
  translate}.
\newblock In \emph{International Conference on Learning Representations}, 2015.

\bibitem[Bao et~al.(2022)Bao, Li, Zhu, and Zhang]{bao2022analyticdpm}
Bao, F., Li, C., Zhu, J., and Zhang, B.
\newblock Analytic-{DPM}: An analytic estimate of the optimal reverse variance
  in diffusion probabilistic models.
\newblock In \emph{International Conference on Learning Representations}, 2022.

\bibitem[Bengio et~al.(2015)Bengio, Vinyals, Jaitly, and
  Shazeer]{bengio2015scheduled}
Bengio, S., Vinyals, O., Jaitly, N., and Shazeer, N.
\newblock Scheduled sampling for sequence prediction with recurrent neural
  networks.
\newblock \emph{Neural Information Processing Systems}, 28, 2015.

\bibitem[Benny \& Wolf(2022)Benny and Wolf]{benny2022dynamic}
Benny, Y. and Wolf, L.
\newblock Dynamic dual-output diffusion models.
\newblock In \emph{Computer Vision and Pattern Recognition}, 2022.

\bibitem[Chen et~al.(2020)Chen, Zhang, Zen, Weiss, Norouzi, and
  Chan]{chen2020wavegrad}
Chen, N., Zhang, Y., Zen, H., Weiss, R.~J., Norouzi, M., and Chan, W.
\newblock {WaveGrad}: Estimating gradients for waveform generation.
\newblock In \emph{International Conference on Learning Representations}, 2020.

\bibitem[Choi et~al.(2021)Choi, Kim, Jeong, Gwon, and Yoon]{choi2021ilvr}
Choi, J., Kim, S., Jeong, Y., Gwon, Y., and Yoon, S.
\newblock {ILVR}: Conditioning method for denoising diffusion probabilistic
  models.
\newblock In \emph{International Conference on Computer Vision}, 2021.

\bibitem[Elfwing et~al.(2018)Elfwing, Uchibe, and Doya]{silu}
Elfwing, S., Uchibe, E., and Doya, K.
\newblock Sigmoid-weighted linear units for neural network function
  approximation in reinforcement learning.
\newblock \emph{Neural Networks}, 107:\penalty0 3--11, 2018.

\bibitem[Elliott et~al.(1996)Elliott, Rothenberg, and
  Stock]{elliott1996efficient}
Elliott, G., Rothenberg, T.~J., and Stock, J.~H.
\newblock Efficient tests for an autoregressive unit root.
\newblock \emph{Econometrica}, pp.\  813--836, 1996.

\bibitem[Fan et~al.(2022)Fan, Zheng, Yi, Cao, Fu, Bian, and Liu]{fan2022depts}
Fan, W., Zheng, S., Yi, X., Cao, W., Fu, Y., Bian, J., and Liu, T.-Y.
\newblock {DEPTS}: Deep expansion learning for periodic time series
  forecasting.
\newblock In \emph{International Conference on Learning Representations}, 2022.

\bibitem[Gong et~al.(2023)Gong, Li, Feng, Wu, and Kong]{DiffuSeq}
Gong, S., Li, M., Feng, J., Wu, Z., and Kong, L.
\newblock {DiffuSeq}: Sequence to sequence text generation with diffusion
  models.
\newblock In \emph{International Conference on Learning Representations}, 2023.

\bibitem[Graves et~al.(2013)Graves, Jaitly, and Mohamed]{graves2013hybrid}
Graves, A., Jaitly, N., and Mohamed, A.-R.
\newblock Hybrid speech recognition with deep bidirectional {LSTM}.
\newblock In \emph{IEEE Workshop on Automatic Speech Recognition and
  Understanding}, pp.\  273--278, 2013.

\bibitem[Henrique et~al.(2019)Henrique, Sobreiro, and
  Kimura]{henrique2019literature}
Henrique, B.~M., Sobreiro, V.~A., and Kimura, H.
\newblock Literature review: Machine learning techniques applied to financial
  market prediction.
\newblock \emph{Expert Systems with Applications}, 124:\penalty0 226--251,
  2019.

\bibitem[Ho et~al.(2020)Ho, Jain, and Abbeel]{ddpm20}
Ho, J., Jain, A., and Abbeel, P.
\newblock Denoising diffusion probabilistic models.
\newblock In \emph{Neural Information Processing Systems}, 2020.

\bibitem[Hochreiter \& Schmidhuber(1997)Hochreiter and
  Schmidhuber]{hochreiter1997long}
Hochreiter, S. and Schmidhuber, J.
\newblock Long short-term memory.
\newblock \emph{Neural Computation}, 9\penalty0 (8):\penalty0 1735--1780, 1997.

\bibitem[Kim et~al.(2022)Kim, Kim, and Yoon]{kim2022guided}
Kim, H., Kim, S., and Yoon, S.
\newblock Guided-{TTS}: A diffusion model for text-to-speech via classifier
  guidance.
\newblock In \emph{International Conference on Machine Learning}, 2022.

\bibitem[Kim et~al.(2021)Kim, Kim, Tae, Park, Choi, and
  Choo]{kim2021reversible}
Kim, T., Kim, J., Tae, Y., Park, C., Choi, J.-H., and Choo, J.
\newblock Reversible instance normalization for accurate time-series
  forecasting against distribution shift.
\newblock In \emph{International Conference on Learning Representations}, 2021.

\bibitem[Kingma \& Ba(2015)Kingma and Ba]{kingma2014adam}
Kingma, D.~P. and Ba, J.
\newblock Adam: A method for stochastic optimization.
\newblock In \emph{International Conference on Learning Representations}, 2015.

\bibitem[Kingma \& Welling(2014)Kingma and Welling]{kingma2013auto}
Kingma, D.~P. and Welling, M.
\newblock Auto-encoding variational {B}ayes.
\newblock In \emph{International Conference on Learning Representations}, 2014.

\bibitem[Kong et~al.(2020)Kong, Ping, Huang, Zhao, and
  Catanzaro]{kong2020diffwave}
Kong, Z., Ping, W., Huang, J., Zhao, K., and Catanzaro, B.
\newblock {DiffWave}: A versatile diffusion model for audio synthesis.
\newblock In \emph{International Conference on Learning Representations}, 2020.

\bibitem[Lai et~al.(2018)Lai, Chang, Yang, and Liu]{lai2018modeling}
Lai, G., Chang, W.-C., Yang, Y., and Liu, H.
\newblock Modeling long-and short-term temporal patterns with deep neural
  networks.
\newblock In \emph{SIGIR Conference on Research \& Development in Information
  Retrieval}, 2018.

\bibitem[Li et~al.(2019)Li, Jin, Xuan, Zhou, Chen, Wang, and
  Yan]{li2019enhancing}
Li, S., Jin, X., Xuan, Y., Zhou, X., Chen, W., Wang, Y.-X., and Yan, X.
\newblock Enhancing the locality and breaking the memory bottleneck of
  transformer on time series forecasting.
\newblock In \emph{Neural Information Processing Systems}, 2019.

\bibitem[Li et~al.(2022{\natexlab{a}})Li, Thickstun, Gulrajani, Liang, and
  Hashimoto]{NEURIPS2022DiffusionLM}
Li, X., Thickstun, J., Gulrajani, I., Liang, P.~S., and Hashimoto, T.~B.
\newblock {Diffusion-LM} improves controllable text generation.
\newblock In \emph{Neural Information Processing Systems}, 2022{\natexlab{a}}.

\bibitem[Li et~al.(2022{\natexlab{b}})Li, Lu, Wang, and Dou]{ligenerative}
Li, Y., Lu, X., Wang, Y., and Dou, D.
\newblock Generative time series forecasting with diffusion, denoise, and
  disentanglement.
\newblock In \emph{Neural Information Processing Systems}, 2022{\natexlab{b}}.

\bibitem[Liu et~al.(2021)Liu, Yu, Liao, Li, Lin, Liu, and
  Dustdar]{liu2021pyraformer}
Liu, S., Yu, H., Liao, C., Li, J., Lin, W., Liu, A.~X., and Dustdar, S.
\newblock Pyraformer: Low-complexity pyramidal attention for long-range time
  series modeling and forecasting.
\newblock In \emph{International Conference on Learning Representations}, 2021.

\bibitem[Liu et~al.(2022)Liu, Wu, Wang, and Long]{liu2022non}
Liu, Y., Wu, H., Wang, J., and Long, M.
\newblock Non-stationary transformers: Rethinking the stationarity in time
  series forecasting.
\newblock In \emph{Neural Information Processing Systems}, 2022.

\bibitem[Lu et~al.(2022)Lu, Zhou, Bao, Chen, Li, and Zhu]{lu2022dpmsolver}
Lu, C., Zhou, Y., Bao, F., Chen, J., Li, C., and Zhu, J.
\newblock {DPM-Solver}: A fast {ODE} solver for diffusion probabilistic model
  sampling in around 10 steps.
\newblock In \emph{Neural Information Processing Systems}, 2022.

\bibitem[Lugmayr et~al.(2022)Lugmayr, Danelljan, Romero, Yu, Timofte, and
  Van~Gool]{lugmayr2022repaint}
Lugmayr, A., Danelljan, M., Romero, A., Yu, F., Timofte, R., and Van~Gool, L.
\newblock Repaint: Inpainting using denoising diffusion probabilistic models.
\newblock In \emph{IEEE/CVF Conference on Computer Vision and Pattern
  Recognition}, 2022.

\bibitem[Nie et~al.(2023)Nie, Nguyen, Sinthong, and Kalagnanam]{nie2022time}
Nie, Y., Nguyen, N.~H., Sinthong, P., and Kalagnanam, J.
\newblock A time series is worth 64 words: Long-term forecasting with
  transformers.
\newblock In \emph{International Conference on Learning Representations}, 2023.

\bibitem[Oreshkin et~al.(2019)Oreshkin, Carpov, Chapados, and
  Bengio]{oreshkin2019n}
Oreshkin, B.~N., Carpov, D., Chapados, N., and Bengio, Y.
\newblock {N-BEATS}: Neural basis expansion analysis for interpretable time
  series forecasting.
\newblock In \emph{International Conference on Learning Representations}, 2019.

\bibitem[Rasul et~al.(2021)Rasul, Seward, Schuster, and Vollgraf]{timegrad}
Rasul, K., Seward, C., Schuster, I., and Vollgraf, R.
\newblock Autoregressive denoising diffusion models for multivariate
  probabilistic time series forecasting.
\newblock In \emph{International Conference on Machine Learning}, 2021.

\bibitem[Rombach et~al.(2022)Rombach, Blattmann, Lorenz, Esser, and
  Ommer]{rombach2022high}
Rombach, R., Blattmann, A., Lorenz, D., Esser, P., and Ommer, B.
\newblock High-resolution image synthesis with latent diffusion models.
\newblock In \emph{IEEE/CVF Conference on Computer Vision and Pattern
  Recognition}, pp.\  10684--10695, 2022.

\bibitem[Sapankevych \& Sankar(2009)Sapankevych and Sankar]{4840324}
Sapankevych, N.~I. and Sankar, R.
\newblock Time series prediction using support vector machines: A survey.
\newblock \emph{IEEE Computational Intelligence Magazine}, 4\penalty0
  (2):\penalty0 24--38, 2009.

\bibitem[Song et~al.(2021)Song, Meng, and Ermon]{song2021denoising}
Song, J., Meng, C., and Ermon, S.
\newblock Denoising diffusion implicit models.
\newblock In \emph{International Conference on Learning Representations}, 2021.

\bibitem[Tashiro et~al.(2021)Tashiro, Song, Song, and Ermon]{tashiro2021csdi}
Tashiro, Y., Song, J., Song, Y., and Ermon, S.
\newblock {CSDI}: Conditional score-based diffusion models for probabilistic
  time series imputation.
\newblock In \emph{Neural Information Processing Systems}, 2021.

\bibitem[Vaswani et~al.(2017)Vaswani, Shazeer, Parmar, Uszkoreit, Jones, Gomez,
  Kaiser, and Polosukhin]{vaswani2017attention}
Vaswani, A., Shazeer, N., Parmar, N., Uszkoreit, J., Jones, L., Gomez, A.~N.,
  Kaiser, {\L}., and Polosukhin, I.
\newblock Attention is all you need.
\newblock \emph{Neural Information Processing Systems}, 30, 2017.

\bibitem[Wang et~al.(2011)Wang, Guo, and Huang]{wang2011review}
Wang, X., Guo, P., and Huang, X.
\newblock A review of wind power forecasting models.
\newblock \emph{Energy procedia}, 12:\penalty0 770--778, 2011.

\bibitem[Wang et~al.(2017)Wang, Yan, and Oates]{wang2017time}
Wang, Z., Yan, W., and Oates, T.
\newblock Time series classification from scratch with deep neural networks: A
  strong baseline.
\newblock In \emph{International Joint Conference on Neural Networks}, 2017.

\bibitem[Williams \& Zipser(1989)Williams and
  Zipser]{williams1989teacherforcing}
Williams, R.~J. and Zipser, D.
\newblock A learning algorithm for continually running fully recurrent neural
  networks.
\newblock \emph{Neural Computation}, 1\penalty0 (2):\penalty0 270--280, 1989.

\bibitem[Wu et~al.(2021)Wu, Xu, Wang, and Long]{wu2021autoformer}
Wu, H., Xu, J., Wang, J., and Long, M.
\newblock Autoformer: Decomposition transformers with auto-correlation for
  long-term series forecasting.
\newblock In \emph{Neural Information Processing Systems}, 2021.

\bibitem[Yang et~al.(2022)Yang, Srivastava, and Mandt]{yang2022diffusion}
Yang, R., Srivastava, P., and Mandt, S.
\newblock Diffusion probabilistic modeling for video generation.
\newblock Technical report, arXiv, 2022.

\bibitem[Zeng et~al.(2023)Zeng, Chen, Zhang, and Xu]{zeng2022transformers}
Zeng, A., Chen, M., Zhang, L., and Xu, Q.
\newblock Are transformers effective for time series forecasting?
\newblock In \emph{AAAI Conference on Artificial Intelligence}, 2023.

\bibitem[Zerveas et~al.(2021)Zerveas, Jayaraman, Patel, Bhamidipaty, and
  Eickhoff]{transformer_repr}
Zerveas, G., Jayaraman, S., Patel, D., Bhamidipaty, A., and Eickhoff, C.
\newblock A transformer-based framework for multivariate time series
  representation learning.
\newblock In \emph{SIGKDD Conference on Knowledge Discovery \& Data Mining},
  2021.

\bibitem[Zhang et~al.(2018)Zhang, Cisse, Dauphin, and
  Lopez-Paz]{zhang2018mixup}
Zhang, H., Cisse, M., Dauphin, Y.~N., and Lopez-Paz, D.
\newblock mixup: Beyond empirical risk minimization.
\newblock In \emph{International Conference on Learning Representations}, 2018.

\bibitem[Zhang \& Yan(2023)Zhang and Yan]{zhang2023crossformer}
Zhang, Y. and Yan, J.
\newblock Crossformer: Transformer utilizing cross-dimension dependency for
  multivariate time series forecasting.
\newblock In \emph{International Conference on Learning Representations}, 2023.

\bibitem[Zhou et~al.(2021)Zhou, Zhang, Peng, Zhang, Li, Xiong, and
  Zhang]{zhou2021informer}
Zhou, H., Zhang, S., Peng, J., Zhang, S., Li, J., Xiong, H., and Zhang, W.
\newblock Informer: Beyond efficient transformer for long sequence time-series
  forecasting.
\newblock In \emph{AAAI Conference on Artificial Intelligence}, 2021.

\bibitem[Zhou et~al.(2022{\natexlab{a}})Zhou, Ma, Wen, Sun, Yao, Jin,
  et~al.]{zhou2022film}
Zhou, T., Ma, Z., Wen, Q., Sun, L., Yao, T., Jin, R., et~al.
\newblock {FiLM}: Frequency improved {L}egendre memory model for long-term time
  series forecasting.
\newblock In \emph{Neural Information Processing Systems}, 2022{\natexlab{a}}.

\bibitem[Zhou et~al.(2022{\natexlab{b}})Zhou, Ma, Wen, Wang, Sun, and
  Jin]{22fedformer}
Zhou, T., Ma, Z., Wen, Q., Wang, X., Sun, L., and Jin, R.
\newblock {FED}former: frequency enhanced decomposed transformer for long-term
  series forecasting.
\newblock In \emph{International Conference on Machine Learning},
  2022{\natexlab{b}}.

\end{thebibliography}
\bibliographystyle{icml2023}

\newpage
\appendix
\onecolumn

\begin{appendices}

\section{Time Series Datasets}
\label{vis:raw_data} 

Since different datasets have different sampling interval lengths (see Table \ref{tab:data}), using the same set of prediction horizons
\{96, 192, 336, 720\}
as in \cite{zhou2021informer,wu2021autoformer} for all datasets may not be
appropriate. For example, the \textit{Exchange} dataset contains daily exchange
rates. A prediction horizon of 720 corresponds to predicting two years into the future. Instead, we set the prediction horizon $H$
 to 14, which corresponds to
 2 weeks into the future,
 which is more reasonable. Similarly, we have $H=168$
 for \textit{ETTh1} (corresponding to 1 week into the future), $H=720$
for \textit{NorPool} (corresponding to 1 month into the future), and so on as
shown in
Table \ref{tab:data}.
Note that some papers
also
choose the prediction length based on the dataset's sampling frequency. For
example,
\citeauthor{liu2021pyraformer}
(\citeyear{liu2021pyraformer}) and
\citeauthor{zhang2023crossformer}
(\citeyear{zhang2023crossformer})
also use 168 (instead of 192) for \textit{ETTh1}, \textit{ECL}, and
\textit{Traffic}.

\begin{figure*}[h!]
\centering
\subfigure[\emph{NorPool}.]
{\includegraphics[width=0.45\textwidth]{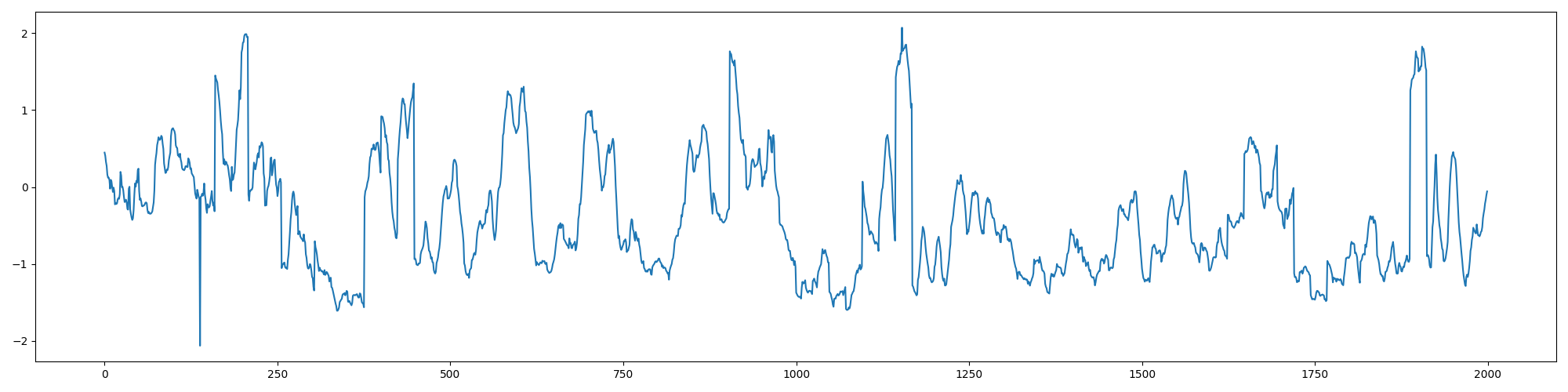}}
\subfigure[\emph{Caiso}.]
{\includegraphics[width=0.45\textwidth]{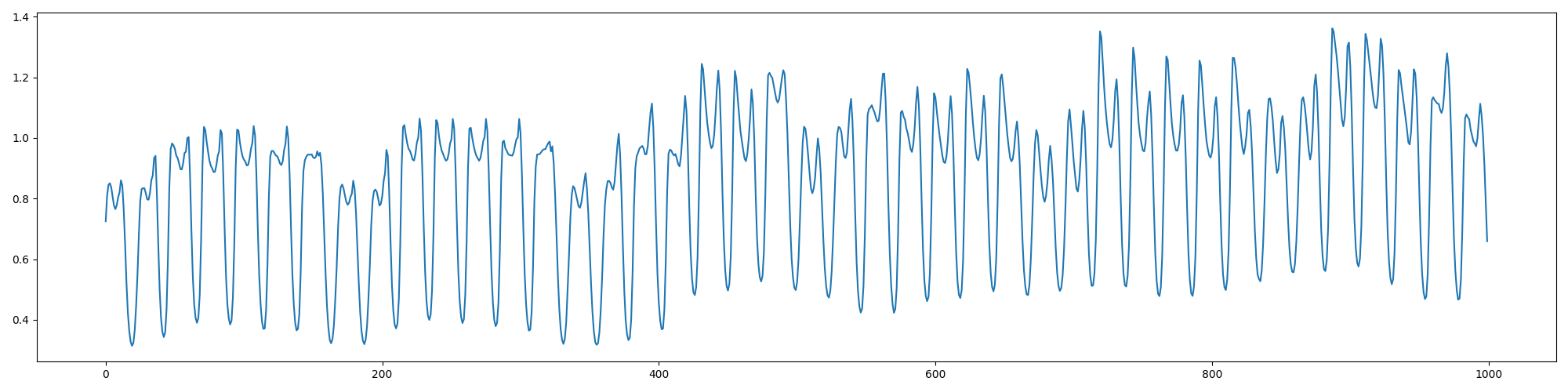}}
\subfigure[\emph{Weather}.]
{\includegraphics[width=0.45\textwidth]
{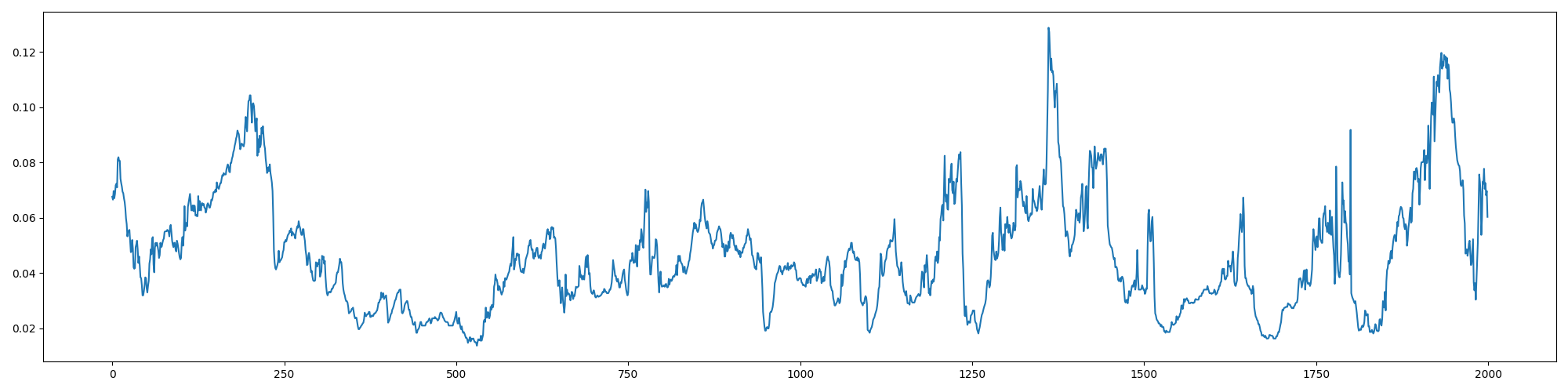}}
\subfigure[\emph{ETTm1}.]
{\includegraphics[width=0.45\textwidth]{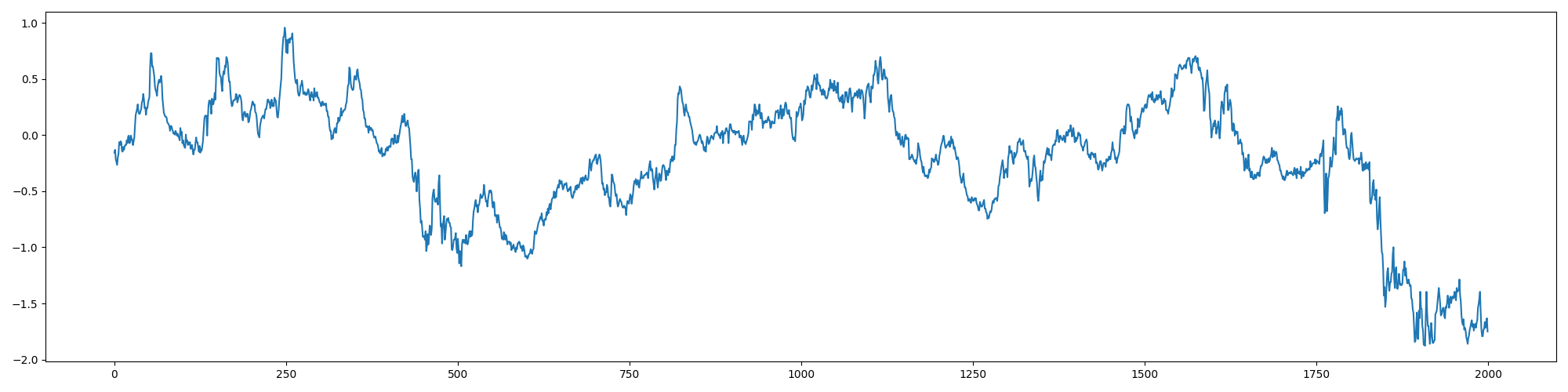}}
\subfigure[\emph{Wind}.]
{\includegraphics[width=0.45\textwidth]{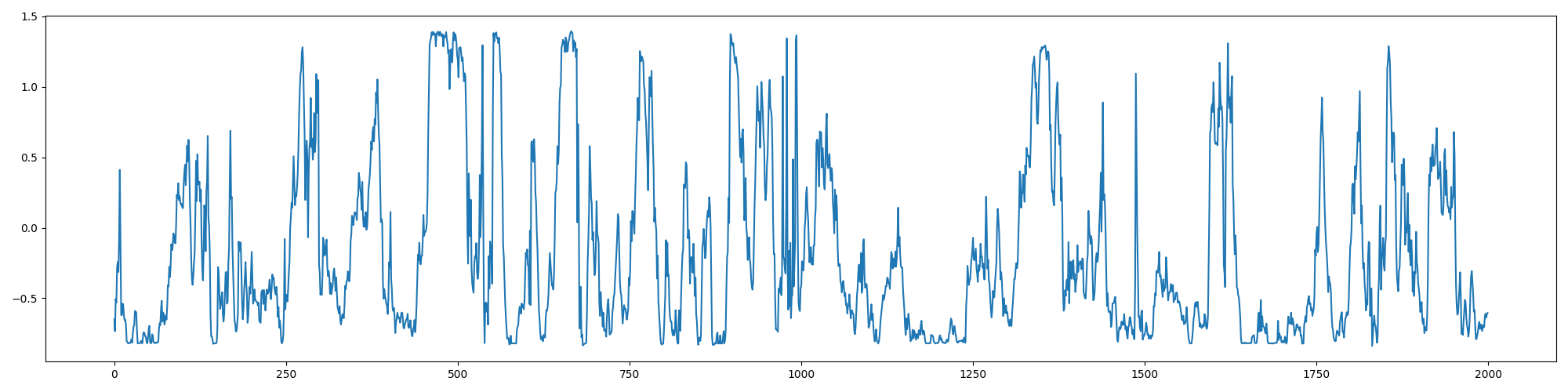}}
\subfigure[\emph{Traffic}.]
{\includegraphics[width=0.45\textwidth]{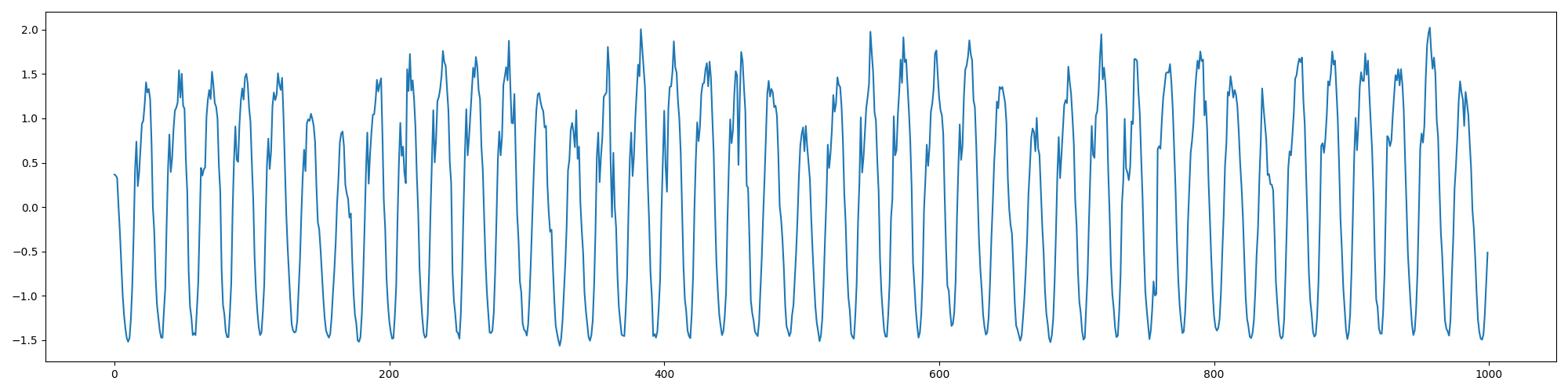}}
\subfigure[\emph{Electricity}.]
{\includegraphics[width=0.45\textwidth]{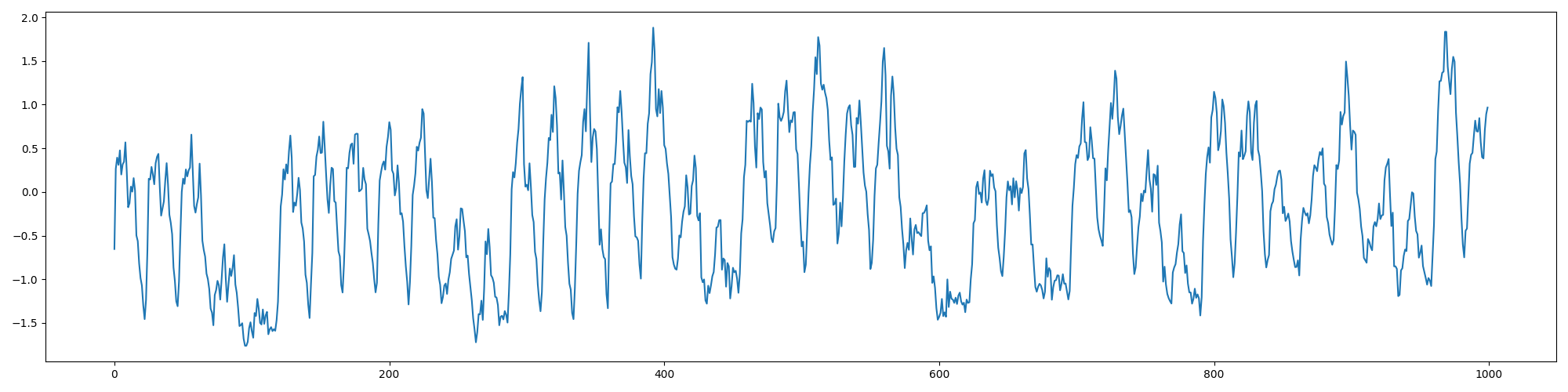}}
\subfigure[\emph{ETTh1}.]
{\includegraphics[width=0.45\textwidth]{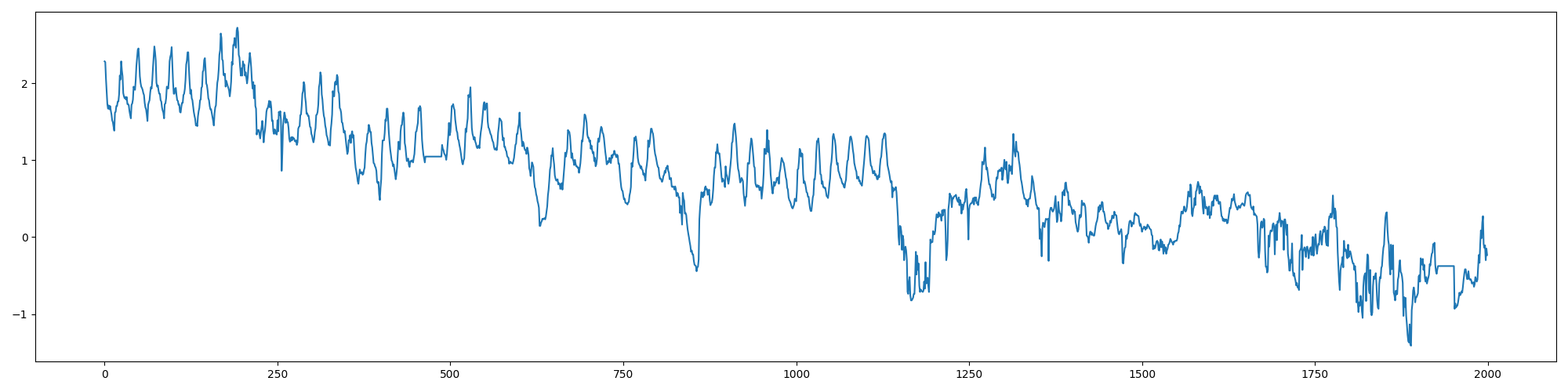}}
\subfigure[\emph{Exchange}.]
{\includegraphics[width=0.45\textwidth]{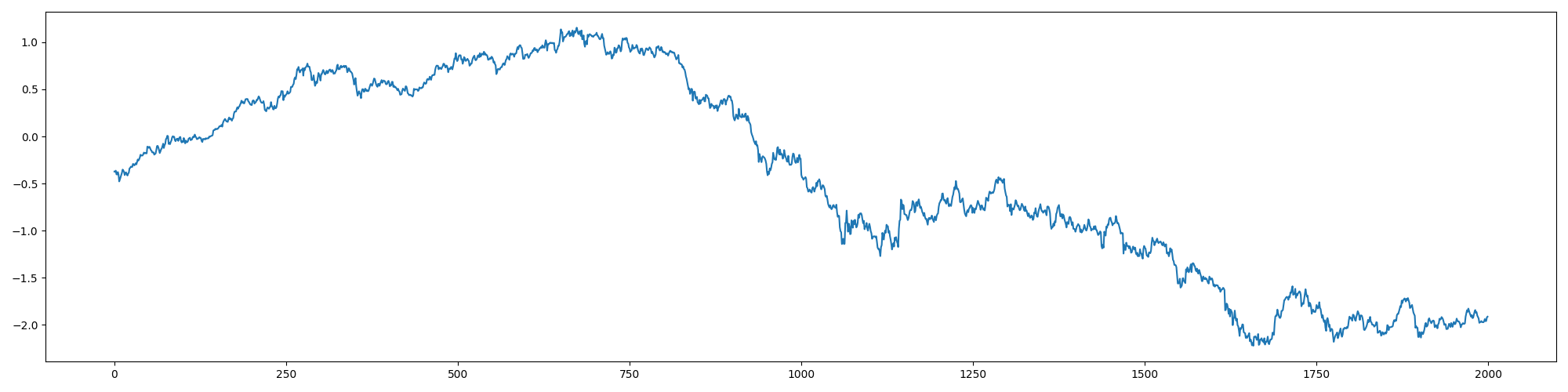}}
\caption{Visualization of the time series datasets.}
\label{fig:data_vis}
\end{figure*}

Figure \ref{fig:data_vis} shows examples of the time series data
used in the experiments.
Since all of them are multivariate, we only show
the last variate.
As can be seen, these datasets have different temporal dynamics.
Moreover, \emph{Caiso}, \emph{Traffic} and \emph{Electricity} show abundant periodic behaviors.

As in \cite{liu2022non},
we use the Augmented Dick-Fuller (ADF) test statistic \cite{elliott1996efficient}
to evaluate if they are non-stationary.
The null hypothesis is that the time series
is not stationary (has some time-dependent structure)
and can be represented by a unit root.
The test statistic results are shown in Table \ref{tab:ADF}.
As can be seen,
with a threshold of 5\%,
\emph{Caiso},
\emph{ETTm1},
\emph{ETTh1},
and
\emph{Exchange}
are  considered
non-stationary.

\begin{table*}[h!]
\centering
\caption{Evaluation  of non-stationarity by the Augmented Dick-Fuller (ADF) test. }
\renewcommand\arraystretch{1.2}
\begin{tabular}{c|ccccccccc}%
\midrule[1pt]
\multirow{1}[0]{*}{}  & \textit{NorPool}& \textit{Caiso} & \textit{Weather}  &  \textit{ETTm1}  &   \textit{Wind} & \textit{Traffic}  & \textit{Electricity} & \textit{ETTh1} & \textit{Exchange} \\\midrule
{ADF statistic} &-6.862&-2.218&-4.371&-1.734&-6.658&-2.801&-2.797&-2.571&-0.001\\
p-value &0&0.2&0&0.414&0&0.005&0.006&0.099&0.095\\
\midrule[1pt]
\end{tabular}
\label{tab:ADF}
\end{table*}

\section{Implementation Details}\label{appendix:Implementation}

\subsection{Network Architecture}

The proposed diffusion model has several subnetworks:
the conditioning network $\gF$,
and
the denoising network's
encoder/decoder.
Table \ref{tab:architecture} shows the subnetwork's input and output sizes,  where $d$ is the number of variables, and $d'=d''= 256$.
Each subnetwork is
constructed by stacking a number of convolutional blocks. 
The configuration of each convolutional block is shown in Table \ref{tab:conv}.

\begin{table*}[h!]
\centering
\caption{Input and output sizes
of the subnetworks.}
\renewcommand\arraystretch{1.2}
\begin{tabular}{cccc}%
\midrule[1pt]
\multirow{1}[0]{*}{}  & input size & output size \\\midrule
$\gF$  & $d\times L$ & $d\times H$\\
encoder  & $2d'\times H$ & $d''\times H$\\
decoder  & $(2d+d")\times L$ & $d\times H$\\
\midrule[1pt]
\end{tabular}
\label{tab:architecture}
\end{table*}

\begin{table*}[h!]
\centering
\caption{Configuration of the convolutional block.}
\renewcommand\arraystretch{1.2}
\begin{tabular}{ccl}%
\midrule[1pt]
\multirow{1}[0]{*}{layer } & operator & default parameters \\\midrule
1  & Conv1d & in channel=256, out channel=256, kernel size=3, stride=1, padding=1 \\
2  & BatchNorm1d & number of features=256\\
3  & LeakyReLU & negative slope=0.1\\
4  & Dropout & dropout rate=0.1\\
\midrule[1pt]
\end{tabular}
\label{tab:conv}
\end{table*}

\subsection{Baselines}
Code for the baselines are downloaded from the following.
  (i) TimeGrad: \url{https://github.com/ForestsKing/TimeGrad};
  (ii) CSDI: \url{https://github.com/ermongroup/CSDI};
  (iii) SSSD: \url{https://github.com/AI4HealthUOL/SSSD};
  (iv) D$^3$VAE: \url{https://github.com/ramber1836/d3vae};
  (v) FiLM: \url{https://github.com/DAMO-DI-ML/NeurIPS2022-FiLM};
  (vi) Depts: \url{https://github.com/weifantt/DEPTS};
  (vii) NBeats: \url{https://github.com/ServiceNow/N-BEATS};
  (viii) PatchTST: \url{https://github.com/yuqinie98/PatchTST/tree/main/PatchTST_self_supervised};
  (ix) Fedformer: \url{https://github.com/DAMO-DI-ML/ICML2022-FEDformer};
  (x) Autoformer: \url{https://github.com/thuml/Autoformer};
  (xi) Pyraformer: \url{https://github.com/ant-research/Pyraformer};
  (xii) Informer: \url{https://github.com/zhouhaoyi/Informer2020};
  (xiii) Transformer: \url{https://github.com/thuml/Autoformer/blob/main/models/Transformer.py};
  (xiv) DLinear: \url{https://github.com/ioannislivieris/DLinear};
  (xv) LSTMa: \url{https://pytorch.org/tutorials/intermediate/seq2seq_translation_tutorial.html}.

\newpage
\section{Using Channel-Independence on Multivariate Time Series Datasets}\label{appendix:channel_independence}

Recall from Section~\ref{sec:setup} that
channel-independence
is not used on PatchTST for the experiments in Section~\ref{sec:expt}.
In this section,
we compare the proposed TimeDiff with
PatchTST, FedFormer, Autoformer and Informer
under
the channel-independence setup.
In other words, each variate of the multivariate time series is predicted
independently. 
Table \ref{tab:channel_ind} shows the testing
MSEs.
As can be seen,
TimeDiff still outperforms the other baselines most
of the time under this setup.

\begin{table}[h!]
\centering
\footnotesize
\caption{Testing MSEs under the channel-independence setup.
Results on PatchTST, FedFormer, Autoformer and Informer are from Table 15 in \cite{nie2022time}. The best is in bold, while the second best is underlined.}
\renewcommand\arraystretch{1.2}
\begin{tabular}{c|cccc|cccc}
\midrule[1pt]
	& \multicolumn{4}{c|}{\textit{ETTh1}}		&
	\multicolumn{4}{c}{\textit{ETTm1}}  \\	
& $H=96$  & 	$H=192$ & 	$H=336$ & 	$H=720$ & 	$H=96$ & 	$H=192$ & 	$H=336$ &
$H=720$\\\midrule
TimeDiff	&  \textbf{0.371}	& \textbf{0.405}& 	\textbf{0.430}& 	\textbf{0.437}& 	\textbf{0.287}& 	\textbf{0.327}& 	\underline{0.368}& 	\textbf{0.414}\\
PatchTST	&\underline{0.375}& 	\underline{0.414}	& \underline{0.431}& 	\underline{0.449}	& \underline{0.290}	& \underline{0.332}& 	\textbf{0.366}& 	\underline{0.420}\\
FedFormer	& 0.387& 	0.439& 	0.479	& 0.485& 	0.408	& 0.445& 	0.476& 	0.533\\
Autoformer	& 0.414& 	0.453	& 0.496	& 0.662& 	0.455& 	0.598	& 0.566	& 0.680\\
Informer	& 0.590& 	0.677& 	0.710& 	0.777& 	0.383& 	0.42& 	0.465& 	0.529\\\midrule
\end{tabular}
\label{tab:channel_ind}
\end{table}

\end{appendices}

\end{document}